\documentclass[10pt,twocolumn,letterpaper]{article}

\usepackage{style}              %

\newcommand{\siso}{SISO}
\usepackage[ruled]{algorithm2e} %

\usepackage{placeins} 

\definecolor{iccvblue}{rgb}{0.21,0.49,0.74}
\usepackage[pagebackref,breaklinks,colorlinks,allcolors=iccvblue]{hyperref}

\title{Single Image Iterative Subject-driven Generation and Editing}

\author{Yair Shpitzer\\
Bar-Ilan University\\
\and
Gal Chechik\\
Bar-Ilan University, NVIDIA\\
\and
Idan Schwartz\\
Bar-Ilan University\\
}

\begin{document}

\twocolumn[{%
\maketitle
\renewcommand\twocolumn[1][]{#1}%
\begin{center}
    \centering
    \includegraphics[width=\textwidth]{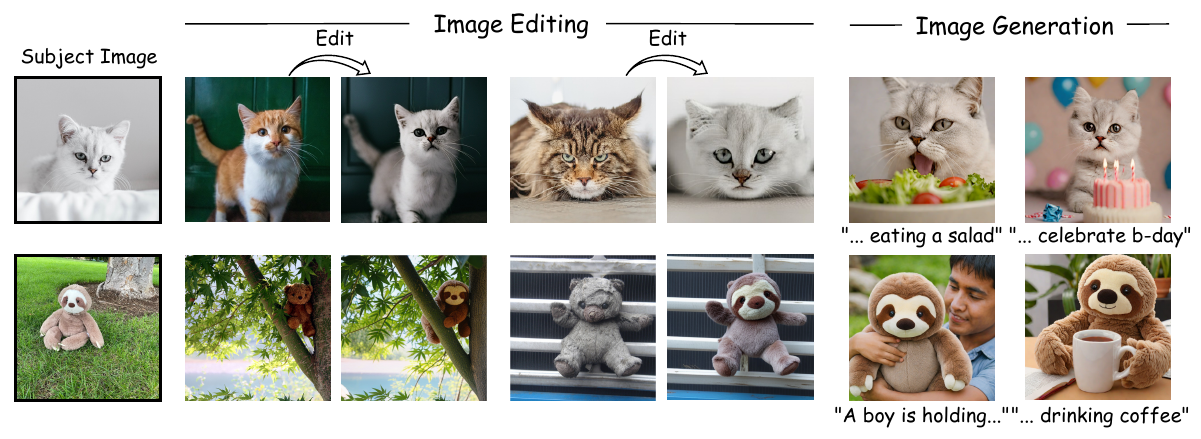}
    \captionsetup{type=figure}\caption{\siso\ is an inference-time optimization method to personalize images \textit{from a single subject image} without training . \siso\ can personalize the subject of a given image or generate new images with the personal subject.}
    \label{fig:teaser}
\end{center}
    }]

\begin{abstract}
Personalized image generation and image editing from an image of a specific subject is at the research frontier. It becomes particularly challenging when one only has a few images of the subject, or even a single image. A common approach to personalization is concept learning, which can integrate the subject into existing models relatively quickly but produces images whose quality tends to deteriorate quickly when the number of subject images is small. Quality can be improved by pre-training an encoder, but training restricts generation to the training distribution, and is time consuming. It is still a difficult and open challenge to personalize image generation and editing from a single image without training.  
Here, we present SISO, a new, training-free approach based on \textit{optimizing a  similarity score with an input subject image}. More specifically, \siso{} iteratively generates images and optimizes the model based on loss of similarity to the given subject image until a satisfactory level of similarity is achieved, allowing plug-and-play optimization to any image generator. We evaluated SISO in two tasks, image editing and image generation, using a diverse data set of personal subjects, and demonstrate significant improvements over existing methods in image quality, subject fidelity, and background preservation.
\end{abstract}

\bibliographystyle{ieeenat_fullname}

\section{Introduction}

Subject-driven text-conditioned image generation and editing combines the ease-of-use of prompt-conditioning with the superior visual control provided when creating visual content using \textit{personalized} elements. It is 
crucial for creative expression, from advertising to digital art, but remains a challenging task when only few images of the personal element are available. 

The most common approach to personalization is \textit{concept learning}, where a pre-trained model is fine-tuned on a few images of a specific concept~\cite{gal2022textual, ruiz2022dreambooth}. While effective when multiple training samples are available, these methods struggle when given only \textit{a single image}, failing to generalize and often overfitting to the specific details of the input. This leads to style leakage and structural distortions rather than accurate personalization.  
\textit{encoder-based methods} \cite{gal2023encoder,li2023blip, ye2023ipadapter} adapt better to a single image by training on a diverse set of concepts. However, this training %
requires significant computational resources and dataset-specific tuning, delaying their public availability. As a result, subject-driven generation and editing remain largely \textit{inaccessible for newer models}, leaving the challenge of efficient single-image personalization unsolved.

To address these challenges, we describe a new method called \siso~(Single Image Subject Optimization) . During generation, \siso~ directly optimizes a subject similarity score between the generated image and a single image.  %
Specifically, we show that by using a similarity score based on DINO~\cite{oquab2023dinov2} and IR features~\cite{shao20221stplacesolutiongoogle}\siso{} excels at capturing identity features and filtering out the background even with a single image. By optimizing this score, our method focuses on preserving the identity of the concept rather than other elements of the scene. 

Employing pre-trained score models for fine-tuning a diffusion model presents significant challenges. Current approaches~\cite{dhariwal2021diffusion,ruiz2022dreambooth}  continue the standard optimization of a diffusion process, which can be viewed as predicting the noise of a given latent. They do not work with pixel-level input because they operate on the latent space.
In contrast with these previous methods, our optimization process iteratively takes as input decoded generated images during inference. We generate an image at each step, compute a similarity loss, and update the parameters.  After each step, we generate a new image and repeat the process until a satisfactory level is achieved.

Our method steers the model at inference time by backpropagating through the diffusion process. With the rise of distilled diffusion models that require as few as one denoising step~\cite{flux2024,podell2023sdxl}, our approach becomes significantly more practical. We further describe how \siso{} can be efficiently applied to standard diffusion processes like Sana~\cite{xie2024sana}, which can be computationally expensive. We describe a two-stage training simplification: first, training in an efficient setup with a low number of denoising steps and simple prompts; then, at inference, applying the optimized model with more denoising steps and varied prompts to enhance output quality.

Fig.~\ref{fig:teaser} demonstrates the effectiveness of \siso, personalizing with a single subject image.  
\siso{} allows for highly natural edits, such as accurately replacing the cat while keeping the original cat's stance. For the plush images, we successfully replaced the subject without altering the background, maintaining a natural pose on the tree. Additionally, our image-generation variant showcases the subject's versatility in various complex prompts.

Beyond improving accuracy and image quality, the test-time optimization approach presented here offers two benefits: (i) it is plug-and-play, meaning both the similarity loss and the generative model can easily be replaced, making it very suitable for the high-paced release cycles of image generators; and (ii) the optimization generates an image at each step, making the optimization process visible and able to stop at each point, enhancing user control.

We ran \siso~with a single subject image for both generating and editing images on the ImageHub benchmark, demonstrating significant improvements in image naturalness while maintaining high fidelity in identity and background preservation. Our human evaluations support these results, showing better prompt alignment and naturalness in image generation, as well as enhanced background preservation and naturalness in image editing. We also provide qualitative results illustrating the significant improvements. 

This work has the following contributions:
\begin{itemize}
    \item[(i)] We propose \siso, a novel inference-time iterative optimization technique that alters the subject of a vanilla image generator using only one reference subject image.
    \item[(ii)] We show that \siso~can be applied to two popular tasks: subject-driven image generation and editing, with minor adaptations to the regularization of penalties.
    \item[(iii)] Our results demonstrate significant improvements in single-image subject-driven personalization, opening up a new thread for research in image personalization that, to our knowledge, has not been explored yet.
\end{itemize}

\section{Related Work}
\paragraph{Concept Learning.} Concepts are typically trained using a small set of up to 20 images. Various fine-tuning techniques have been proposed. Initial attempts used prompt tuning, i.e., learning a token representation~\cite{gal2022textual}, and learning negative prompts as well~\cite{dong2022dreamartist}. The following approach updates the entire model~\cite{ruiz2022dreambooth}. Newer variations learn style and content separately~\cite{shah2025ziplora} or consider multiple concepts~\cite{han2023svdiff}. However, these methods often leak style or fail to learn complex objects needed for subject-driven generation, especially with a limited training image set.

\begin{figure*}[t!]
    \centering
    \includegraphics[width=0.85\linewidth]{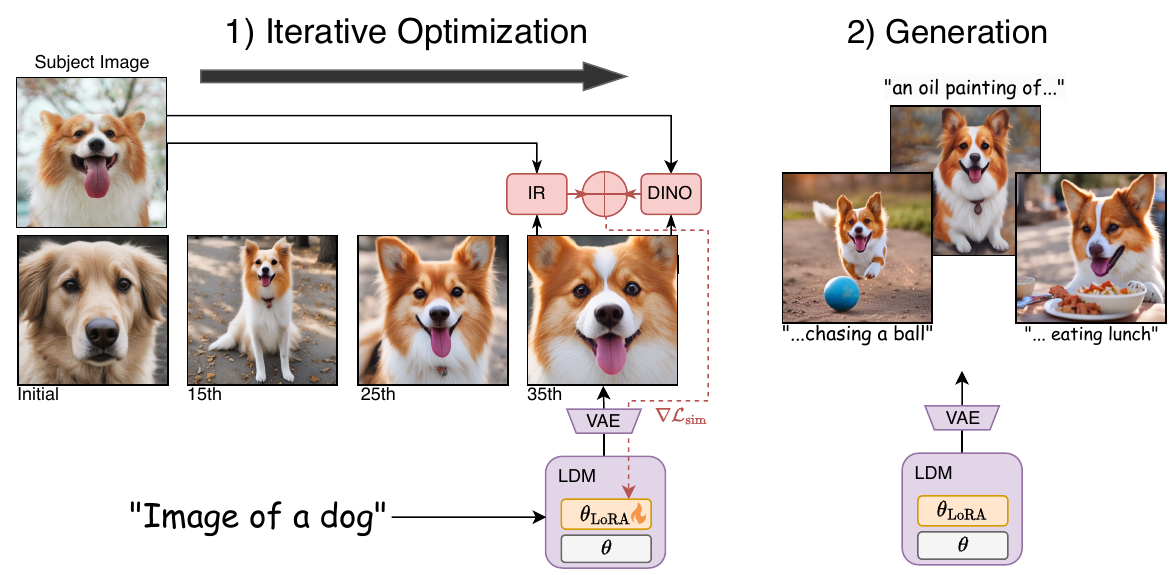}
    \caption{\textbf{\siso\ workflow for image generation}. \siso\ generates images by iteratively optimizing based on pre-trained identity metrics IR and DINO. The added LoRA parameters are updated at each step, while the rest of the models remain frozen. The left panel shows the progress of subject-driven optimization for the prompt ''image of a dog'' by displaying the initial image, followed by the 15th, 25th, and 35th iteration steps. Similarity to the subject image (top) increases during optimization. 
    We find that optimizing with a simple prompt is effective, since the optimized model generates novel images of the subject without further optimization, even with complex prompts, as shown on the right.
    }
    \label{fig:method-generate}
\end{figure*}
\paragraph{Encoder Learning.} 
Early methods trained an encoder to generate an initial subject embedding or to adjust network weights and then fine-tuning during inference for high-quality personalization. However, these methods were often restricted to specific concepts~\cite{gal2023encoder,li2023blip,ruiz2023hyperdreambooth}. Recent approaches studied how to bypass inference-time optimization ~\cite{wei2023diffusion,chen2023subjectdriven,shi2023instantbooth,ye2023ipadapter,jia2023taming,li2023photomaker}. Significant efforts have focused on personalizing human faces, utilizing identity recognition networks or incorporating them as auxiliary losses to enhance identity preservation~\cite{wang2024instantid,yuan2023celebbasis,xiao2023fastcomposer,gal2024lcmlookahead,guo2024pulid,peng2024portraitbooth}. Some recent studies have explored adding cross-attention layers~\cite{gal2024lcmlookahead,guo2024pulid,ye2023ipadapter}. However, methods that encode subjects into existing cross-attention layers tend to preserve the original content more effectively~\cite{li2023photomaker,xiao2023fastcomposer,wang2024moa,alaluf2023neural,tewel2023key,patashnik2025nested}. Despite their advancements, training such encoders still requires substantial computational resources. Recent state-of-the-art encoder solutions, such as the ones proposed for Flux~\cite{flux2024}, require large-scale datasets and extended training times~\cite{tan2024ominicontrol, cai2024diffusion}. Furthermore, to the best of our knowledge, no encoder solution currently exists for Sana~\cite{xie2024sana}. In contrast, our proposed method is plug-and-play, allowing for rapid adaptation to a variety of generative models.

\paragraph{Subject-driven Image Editing.} Initial methods train an adapter to align image encodings with text encodings ~\cite{yang2023paint, song2022objectstitch}. These methods fail on novel concepts. Later methods replaced semantic representations with identity features~\cite{Chen2023AnyDoorZO}. Other works add more control via camera parameters or text prompts~\cite{zhang2023controlcom, yuan2023customnet, xie2023dreaminpainter, pan2024locate}. Following, identity preservation was improved with part-level composition~\cite{Chen2024ZeroshotIE}. Recent works leverage rectified flow models and tailored diffusion noise schedules to enable fast, zero-shot inversion and high-quality semantic image editing~\cite{rout2024semantic,deng2024fireflow,lin2024schedule}.

Another thread explored concept learning from a set of images instead of training an adapter~\cite{gu2023photoswap, li2023dreamedit, gu2024swapanything}. These approaches are closely related to ours; however, learning a concept requires up to 20 images, while \siso~uses a single image. Another recent method is training-free, creating a collage of the reference on the background image~\cite{li2024tuning}. However, this approach is better suited for insertion rather than subject replacement.  Instead, we leverage a subject similarity score that modifies image subjects.

\paragraph{Training Free Image Editing.} Refers to methods with no separate learning phase, commonly used in image editing tasks. Style-transfer methods employ an inversion technique and transfer attention key-and-value representations from a reference style image~\cite{hertz2023style, song2020denoising, huang2017arbitrary, jeong2024visual} or use an encoder~\cite{wang2024instantstyle}. A recent method fuses content and style without inversion~\cite{rout2024rb}. While training-free methods may enable a single image reference for edits, they mostly focus on style. Our approach, which steers the model at inference time, can learn subjects from input reference images.  %

\section{Method}

We introduce \siso{} (Single Image Subject Optimization), a subject-driven conditioning method operating with a \textit{single} subject image. \siso{} 
operates by fine-tuning the diffusion model at inference time, using a loss function computed over the generated image. Specifically, since \siso{} operates over images in pixel space, we can use high-quality pre-trained models that measure object similarity and encourage the model to produce images similar to the desired subject. 
This approach is different from existing approaches that operate by predicting the noise, as done during the training of the diffusion model.

\begin{figure}
    \centering
    \includegraphics[width=1\linewidth]{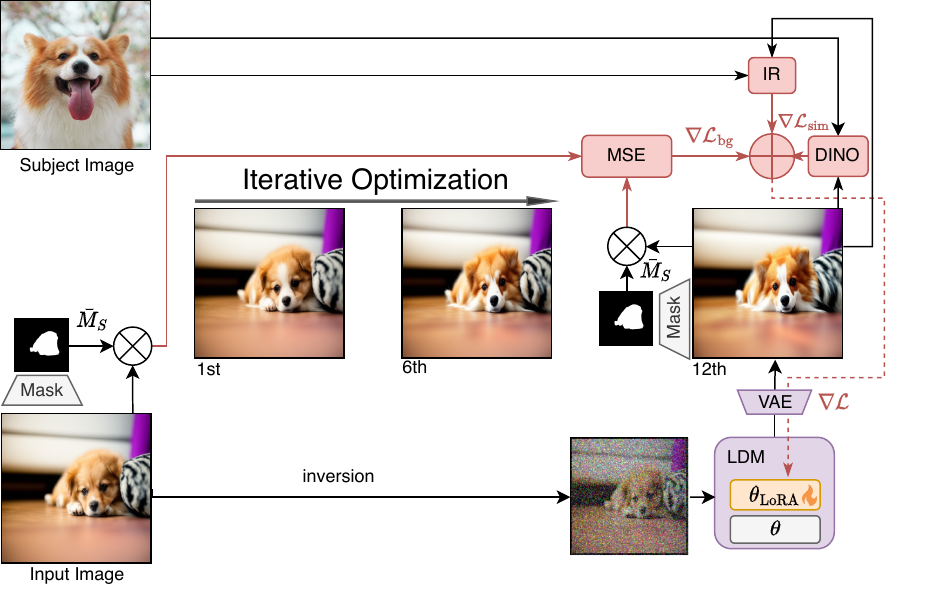}
    \caption{
    \textbf{\siso\ workflow for image editing}. The main differences from %
    generation (Fig. 2) are: (1) %
    Use diffusion inversion to map the input image into a latent begins (bottom); and (2) it adds a background preservation regularization term (Eq. \ref{eq:bg_loss}) 
    }
    \label{fig:method-edit}
\end{figure}

\subsection{Preliminaries: Conditioned Latent Diffusion}%

A conditioned latent diffusion model (LDM) generates an image $x \sim p(x|y)$, where $y$ is the conditioning term, such as text. Training the model is typically achieved  by adding noise to an image and learning to predict the added noise:
$
\min_{\theta} || \hat{\epsilon}_\theta(z_t, t, y) - \epsilon_t ||_2^2
$ . 
Here, $z_t$ is an intermediate noisy latent, $\epsilon_t$ is the added noise up to step $t$, and $\theta$ represents the learnable weights. In many personalization approaches, fine-tuning the model is achieved using the same objective followed during training, namely, reconstruction loss over the latents. In personalization tasks, one is given a set of images of a specific subject that one wishes the model to learn. Here, we assume that only a single subject image is given and denote it by $x_s$. 

\subsection{SISO: Single-Image Subject Optimization}

\siso~optimizes the image generation model during inference using the generated images to compute the loss. By defining the loss in pixel space, we enable using high-quality pre-trained models to measure the similarity between the subject in the generated image and in the input. 

\siso~operates iteratively (Fig. \ref{fig:method-generate} left). We start with randomly initializing low-rank adaptation parameters $\theta_{\text{LoRA}}$ and adding them to a diffusion model following LoRA~\cite{hu2021lora}. We also fix a specific seed for the noise latent $z_T$ and use a deterministic sampler \cite{song2020denoising}. Then, at step $i$ of the iterative process, we generate an image $\hat{x}_i$ using the diffusion model.  The generated image is the output of a differentiable and deterministic LDM, hence any differentiable loss $\mathcal{L}(\theta_{\text{LoRA}}, \hat{x}_i)$ computed over the image can be used for propagating gradients back to the model parameters $\theta_{\text{LoRA}}$.

To preserve subject identity, we set $\mathcal{L}$ to be a subject similarity loss that takes the generated image $\hat{x}_i$ and a reference subject image $x_s$ as input and computes the similarity of subjects across images. We then update the parameters with a gradient descent step: 
\begin{equation} 
    \theta_{\text{LoRA}} \longleftarrow \theta_{\text{LoRA}} + \alpha \frac{\nabla_{\theta_{\text{LoRA}}}\mathcal{L}(\hat{x}_i, x_s)}{|\mathcal{L}(\hat{x}_i, x_s)|^2}. 
\end{equation} 
This update rule is simplified for brevity. In practice, we use Adam optimizer~\cite{kingma2014adam}. After updating the model parameters, we repeat this process iteratively. Since this iterative process involves generating well-formed images, rather than noisy latent, it can be used in an interactive manner. Users can observe and stop the optimization process based on the optimized image displayed at each step, or it can stop automatically using standard early stopping strategies (see Appendix \ref{sec:early_stopping}).

By default, backpropagation through LDM is performed through the entire diffusion process, significantly increasing memory requirements. Our approach is particularly well suited for efficient distilled turbo variants that require only a single diffusion step~\cite{podell2023sdxl}. To support non-distilled models and reduce computational costs, we stop backpropagation after several denoising steps. For instance, with Sana, we backpropagated through the last three denoising steps, which we find sufficient for personalization. This is probably because the final diffusion steps primarily refine local appearance details~\cite{ho2020denoising}.

We now discuss in detail how \siso{} can be used for (i) image generation and (ii) image editing.

\begin{table*}[t]
    \centering
    \caption{Comparison of two baselines for subject-driven image generation using a single reference image per subject. We evaluate identity preservation (DINO, IR), prompt adherence (CLIP-T), and naturalness (FID, KID, CMMD).}
    \resizebox{0.75\linewidth}{!}{%
    \begin{tabular}{l|cc|c|ccc}
    \toprule
        & \multicolumn{2}{c|}{\textbf{Identity Preservation}} & \textbf{Prompt Adherence} & \multicolumn{3}{c}{\textbf{Naturalness}} \\
        & \textbf{DINO$\uparrow$} & \textbf{IR$\uparrow$} & \textbf{CLIP-T$\uparrow$} & \textbf{FID$\downarrow$} & \textbf{KID$\downarrow$} & \textbf{CMMD$\downarrow$} \\
    \midrule
    \textbf{AttnDreamBooth} & 0.47 & 0.51 & 0.29 & 164.4 & 0.004 & 0.41 \\
    \textbf{ClassDiffusion} & \textbf{0.50} & \textbf{0.59} & 0.29 & 166.6 & 0.003 & \textbf{0.18} \\
    \midrule
    \textbf{\siso{} (ours)} & 0.48 & 0.53 & \textbf{0.31} & \textbf{149.2} & \textbf{0.002} & \textbf{0.18} \\
    \bottomrule
    \end{tabular}}
    \label{tab:generation_small}
\end{table*}

\begin{table*}[t]
    \centering
    \caption{Comparing \siso with Dreambooth using three backbone models: SDXL-Turbo, Flux Schnell and Sana. for subject-driven image generation using a single reference image. \siso{} improves prompt adherence while maintaining image fidelity.}
    \resizebox{0.85\textwidth}{!}{%
    \begin{tabular}{lc|cc|c|ccc|c}
    \toprule
        & \textbf{Backbone} & \multicolumn{2}{c|}{\textbf{Identity Preservation}} & \textbf{Prompt Adherence} & \multicolumn{3}{c|}{\textbf{Naturalness}} & \textbf{Diversity} \\
        & & \textbf{DINO$\uparrow$} & \textbf{IR$\uparrow$} & \textbf{CLIP-T$\uparrow$} & \textbf{FID$\downarrow$} & \textbf{KID$\downarrow$} & \textbf{CMMD$\downarrow$} & \textbf{MSE$\uparrow$} \\        
        \midrule
        \textbf{DreamBooth} & SDXL-Turbo & \textbf{0.58} & \textbf{0.67} & 0.28 & 177.69 & 0.010 & 0.85 & 0.05 \\
        \textbf{\siso{} (ours)} & SDXL-Turbo & 0.48 & 0.53 & \textbf{0.31} & \textbf{149.2} & \textbf{0.002} & \textbf{0.18} & \textbf{0.11} \\
        \midrule
        \textbf{DreamBooth} & FLUX Schnell & 0.33 & 0.45 & 0.25 & 227.1 & 0.023 & 1.09 & 0.05 \\
        \textbf{\siso{} (ours)} & FLUX Schnell & \textbf{0.51} & \textbf{0.56} & \textbf{0.31} & \textbf{149.5} & \textbf{0.002} & \textbf{0.14} & \textbf{0.12} \\
        \midrule 
        \textbf{DreamBooth} & Sana & 0.45 & 0.46 & \textbf{0.29} & \textbf{149.5} & \textbf{0.003} & \textbf{0.23} & 0.16 \\
        \textbf{\siso{} (ours)} & Sana & \textbf{0.46} & \textbf{0.51} & \textbf{0.29} & 149.9 & \textbf{0.003} & 0.29 & \textbf{0.19} \\
        \bottomrule
    \end{tabular}}
    \label{tab:generation}
\end{table*}

\subsection{Subject-driven Image Generation}
\label{sec:generation}
To use \siso{} for generation, we expect two inputs: a conditioning prompt and a single reference image of the subject. We define the similarity loss as
\begin{equation}
\mathcal{L}_{\text{sim}}(\hat x_i, x_s) = a\cdot\delta_{\text{DINO}}(\hat x_i, x_s) + b\cdot\delta_{\text{IR}}(\hat x_i, x_s), 
\end{equation}
where $\hat x_i$ is the generated image at optimization step $i$, $\delta_{\text{DINO}}$ and $\delta_{\text{IR}}$ are distances in DINO~\cite{oquab2023dinov2} and IR~\cite{shao20221stplacesolutiongoogle} embedding spaces, and $a,b\in\mathbb{R}$ are calibration hyper-parameters. IR and DINO are suited for assessing the identity distance of objects independent of background influences. Using two metrics in our loss function serves two purposes. (i) They enhance performance thanks to an ``ensemble" effect; and (ii) they serve as a form of penalty regularization, mitigating the risk of mode collapse that might occur when optimizing based on a single metric.

\paragraph{Training Simplification.} To enhance training stability, we find generating simple images using a simple prompt beneficial, as similarity metrics often struggle in complex scenes. Additionally, we observe that training with a low number of denoising steps, even a single step, is sufficient for efficiency.  

Notably, the optimized LoRA weights, even when trained with a simple prompt and minimal denoising steps, can be used for inference with different prompts and more denoising steps to enhance quality.  

This insight inspired a two-stage approach for handling detailed scenes: (1) first, optimize with a simple prompt and a low number of denoising steps, then (2) use the fine-tuned model to generate images with more complex prompts and additional denoising steps. As shown in Fig.~\ref{fig:method-generate} (right), after optimizing LoRA weights for the prompt ``image of a dog,'' the learned subject can be generated for various prompts without further optimization.

\subsection{Subject-driven Image Editing} 

In subject-driven image editing, the model swaps the subject of a given image $\tilde x_0$  with reference image $x_s$ while crucially preserving the background, unlike in image generation, where background coherence with the prompt suffices. Additionally,  editing an image requires converting it into the domain of the diffusion model (see Fig.~\ref{fig:method-edit}).

We begin with inversion using ReNoise inversion~\cite{garibi2024renoise}, which yields faithful inversions (more details in section \ref{sec:diffusion_inversion} of the Appendix). Let $\hat{x}_0 = \operatorname{Inversion}(\tilde x_0)$ be the inverted image of $\tilde x_0$. To preserve the background, we first generate a subject mask $M_s$ by classifying the image $x_s$ and employing object detection with Grounding DINO to identify objects of the same class~\cite{liu2025grounding}. We then extract a segmentation mask from the detected bounding box using SAM~\cite{kirillov2023segment}. The background loss is defined as follows:
\begin{equation} 
    \mathcal{L}_{\text{bg}}(\bar{x}_i, x_s, \hat{x}_0)  = \operatorname{MSE}(\bar{M}_s(\bar{x}_i), \bar{M}_s(\hat{x}_0)),
    \label{eq:bg_loss}
\end{equation} 
where $\bar{M}_s$ is the inverse subject mask, i.e., the subject's background. Intuitively, this loss acts as a penalty for maintaining the background of the original image $\tilde x_0$. Overall, the loss for subject-driven image editing is: 
\begin{equation} 
    \mathcal{L}(\bar{x}_i, x_s, \hat{x}_0)  = \mathcal{L}_{\text{sim}}(\bar{x}_i, x_s)  + c \cdot \mathcal{L}_{\text{bg}}(\bar{x}_i, x_s, \hat{x}_0), 
\end{equation} 
where $c$ is a hyperparameter. We optimize the loss with our iterative inference-time optimization technique.

\section{Experiments}
\label{sec:sim}

\paragraph{Benchmark Dataset and evaluation  protocol.}
We use the benchmark dataset and the experimental protocol from ImagenHub~\cite{ku2023imagenhub}. For subject-driven image \textbf{editing}, their  setup consists of 154 samples, each featuring one of 22 unique subjects from various categories. These include as animals (cat, dog) and day-to-day objects like a backpack, sunglasses, or a teapot. Subject images were taken from  DreamBooth~\cite{ruiz2022dreambooth}. For  subject-driven image \textbf{generation}, the  setup comprises of 150  prompts with 29 unique sample subjects with similar categories.

\paragraph{Implementation details.} For image generation, we used SDXL-Turbo~\cite{sauer2025adversarial}, the distilled version of SDXL~\cite{podell2023sdxl}.  For image editing, we used SD-Turbo\footnote{\url{https://huggingface.co/stabilityai/sd-turbo}}, a distilled version of Stable Diffusion 2.1~\cite{rombach_2022_cvpr}. We set the loss calibration hyperparameters to $a=1, b=1, c=10$, and the learning rate to $\alpha=3e^{-4}$. The resolution in all our experiments is $512\times512$.

\paragraph{Baselines.} 
Since our task is to efficiently adapt a pre-trained image generator using a \textit{single image} of a reference subject, we 
compare \siso{} against baselines that can operate without requiring to train an  encoder learning. %
For image generation, we compared with AttnDreamBooth~\cite{pang2025attndreambooth}. It improves over DreamBooth~\cite{ruiz2022dreambooth} with a three-stage process, optimizing textual embedding, cross-attention layers, and the U-Net. We also compared with ClassDiffusion, which uses a semantic preservation loss~\cite{huang2024classdiffusion}. For image editing, we used SwapAnything, which employs masked latent blending and appearance adaptation~\cite{gu2024swapanything}. All the methods above use concept learning to depict the subject and typically require up to 20 subject images for accurate performance. However, here, we use them with a single image. We also compared with TIGIC, a training-free technique that uses an attention-blending strategy during denoising~\cite{li2024tuning}.

\subsection{Evaluation Metrics}

\paragraph{Identity Preservation.}  
To evaluate subject similarity, we crop the subject using Grounding DINO~\cite{liu2025grounding} and compare it using: (i) DINO distance for instance similarity, particularly for animals, (ii) IR features, effective in item similarity~\cite{shao20221stplacesolutiongoogle}, and (iii) CLIP-I, which measures class-level similarity~\cite{radford2021learning}.  

\noindent\newline\textbf{Naturalness.}  
To assess image realism, we compare generated images against a reference set: vanilla Stable Diffusion outputs for generation and input images for editing. We compute three metrics: FID~\cite{heusel2017gans}, KID~\cite{binkowski2018demystifying}, which has been shown to be more stable in small datasets, and CMMD~\cite{jayasumana2024rethinking} for semantically richer CLIP-based evaluation.

\begin{table}[t!]
    \caption{Subject-driven image editing. All experiments used a single reference image per subject. We report identity preservation (DINO, IR, CLIP-I), background preservation (LPIPS), and naturalness (FID, KID, CMMD).}
    \centering
    \setlength{\tabcolsep}{2pt}
    \resizebox{\linewidth}{!}{%
    \begin{tabular}{lccc|c|ccc}
    \toprule
        \textbf{} & \multicolumn{3}{c|}{\textbf{Identity Preservation}} & \textbf{Background } & \multicolumn{3}{c}{\textbf{Naturalness}} \\

         &  & & \textbf{} & \textbf{Preservation} &  &  &  \\

        \textbf{} & \textbf{DINO $\uparrow$} & \textbf{IR$\uparrow$} & \textbf{CLIP-I} & \textbf{LPIPS $\downarrow$} & \textbf{FID $\downarrow$} & \textbf{KID $\downarrow$} & \textbf{CMMD $\downarrow$} \\
        \midrule
        \textbf{TIGIC} & 0.51 & 0.58 & 0.77 & 0.22 & 143.26 & 0.0066 & 0.759 \\
        \textbf{SwapAnything} & 0.45 & 0.60 & 0.74 & \textbf{0.11} & 185.74 & 0.0277 & 1.101 \\
        \textbf{\siso{} (ours)} & \textbf{0.55} & \textbf{0.75} & \textbf{0.80} & 0.14 & \textbf{114.83} & \textbf{0.0031} & \textbf{0.475} \\
        \bottomrule
    \end{tabular}}
    \label{tab:edit}
\end{table}

\begin{table}[t!]
    \caption{Ablation %
        for image generation.
        We report identity preservation (DINO, IR) and prompt adherence (CLIP-T)}
    \centering
\resizebox{\linewidth}{!}{%
    \begin{tabular}{lcc|c}
    \toprule
        \textbf{} & \multicolumn{2}{c|}{\textbf{Identity Preservation}} & \textbf{Prompt Adherence} \\
        \textbf{} & \textbf{DINO $\uparrow$} & \textbf{IR$\uparrow$}  & \textbf{CLIP-T $\uparrow$} \\
        \midrule
        \textbf{\siso{} (ours)} & 0.48 & 0.53 & \textbf{0.31} \\
        
        \textbf{Ours - w/o Prompt Simpl.} & \textbf{0.52} & \textbf{0.62} & 0.29 \\
        \textbf{Ours - w/o DINO} & 0.44 & 0.50 & \textbf{0.31} \\
        \textbf{Ours - w/o IR} & 0.49 & 0.50 & \textbf{0.31} \\
        \bottomrule
    \end{tabular}}
    \label{tab:ablation-gen}
\end{table}

\begin{table}[t!]
    \caption{Ablation for image editing. We report identity preservation (DINO, IR, CLIP-I) and background preservation (LPIPS)}
    \centering
\resizebox{\linewidth}{!}{%
    \begin{tabular}{lccc|c}
    \toprule
        \textbf{} & \multicolumn{3}{c|}{\textbf{Identity Preservation}} & \textbf{BG Preservation} \\
        \textbf{} & \textbf{DINO $\uparrow$} & \textbf{IR$\uparrow$} & \textbf{CLIP-I$\uparrow$} & \textbf{LPIPS $\downarrow$} \\
        \midrule
        \textbf{\siso{} (ours)} & \textbf{0.55} & 0.75 & \textbf{0.80} & 0.14 \\
        \midrule
        \textbf{Ours - w/o BG Pres.} & 0.55 & \textbf{0.76 }& 0.80 & 0.18 \\
        \textbf{Ours - w/o DINO} & 0.49 & 0.74 & 0.78 & 0.13 \\
        \textbf{Ours - w/o IR} & 0.54 & 0.56 & 0.78 & \textbf{0.12} \\
        \bottomrule
    \end{tabular}}
    \label{tab:ablation-edit}
\end{table}

\begin{table}[t!]
    \caption{User study for image editing (left) and generation (right). values are the win rate of our method (fraction of preferred cases) against the leading baseline. $\pm$ denotes the standard error of the mean (SEM) based on a binomial distribution.}
    \centering
    \scalebox{0.9}{
    \begin{tabular}{lc cc}
    \toprule
    & \textbf{TIGIC } & \textbf{ClassDiffusion } \\
    & \textbf{(Editing)} & \textbf{(Generation)} \\
    \midrule
    \textbf{Identity Preservation} & 0.45 $\pm$ 0.05 & 0.47 $\pm$ 0.05 \\
    \textbf{Naturalness} & \textbf{0.58 $\pm$ 0.05} & \textbf{0.65 $\pm$ 0.05} \\
    \textbf{Background Preservation} & \textbf{0.60 $\pm$ 0.05} & - \\
    \textbf{Prompt Adherence} & - & \textbf{0.69 $\pm$ 0.05} \\
    \bottomrule
    \end{tabular}}
    \label{tab:human}
\end{table}

\begin{figure*}
    \centering
    \setlength{\tabcolsep}{1pt}
    \begin{tabular}{ccc cccccc }
        & & Subject Image & & Ours & ClassDiffusion & AttnDreamBooth & DreamBooth & TI \\

        \raisebox{26pt}{\rotatebox[origin=t]{90}{\small }} &
        \raisebox{26pt}{\rotatebox[origin=t]{90}{\small ... in Paris}} &
        \includegraphics[width=0.14\linewidth]{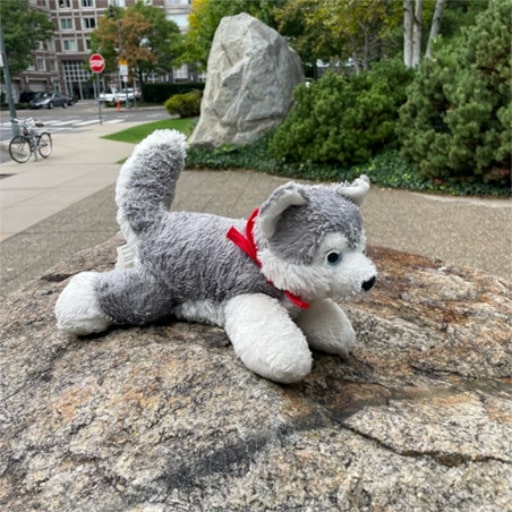} &
                \raisebox{28pt}{$\rightarrow$} &
        \includegraphics[width=0.14\linewidth]{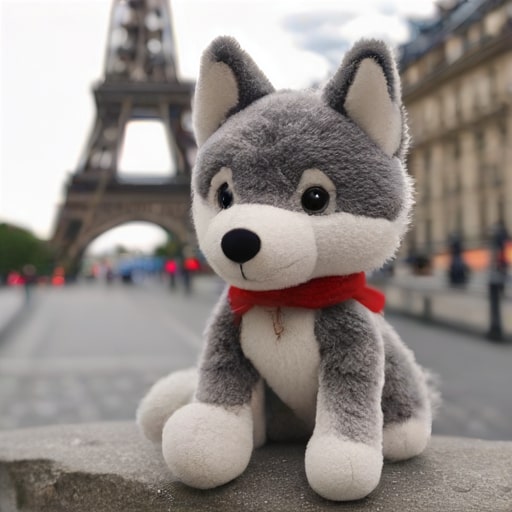} &
        \includegraphics[width=0.14\linewidth]{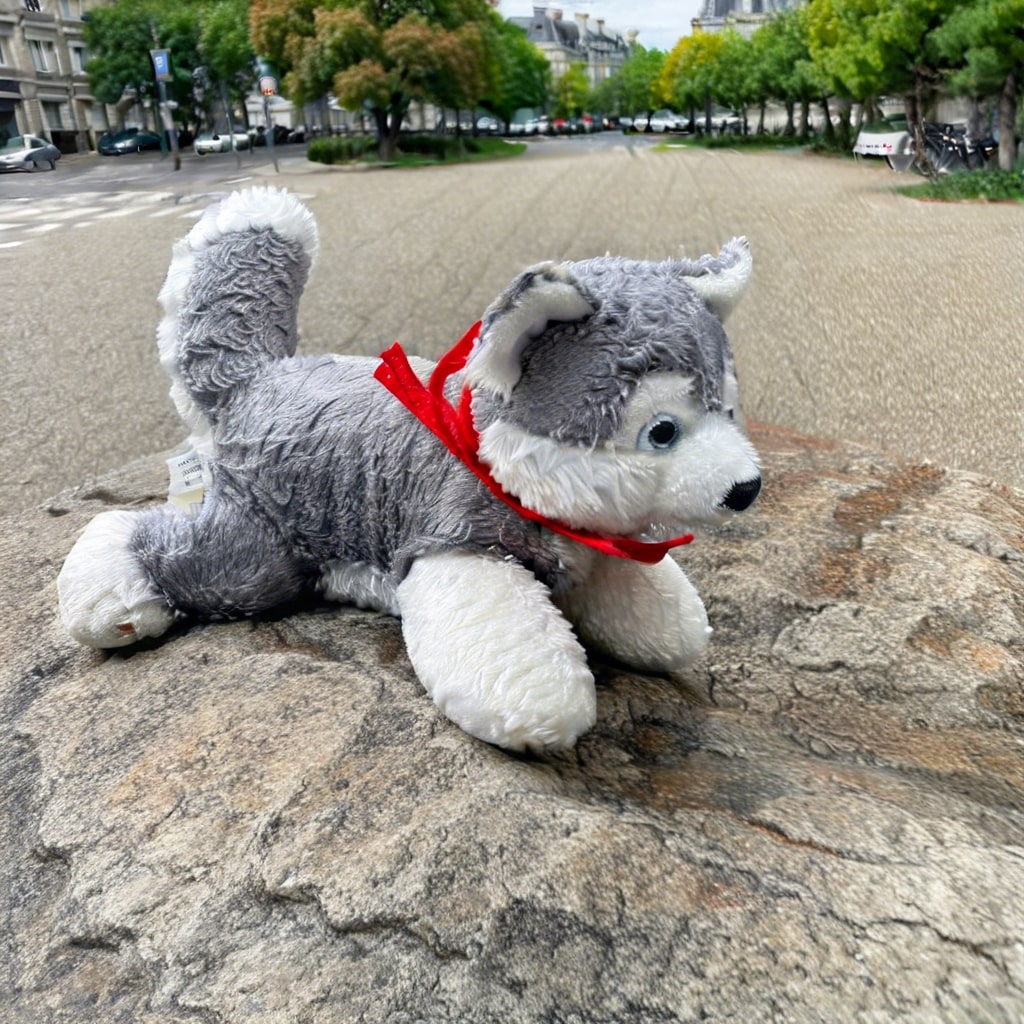} &
        \includegraphics[width=0.14\linewidth]{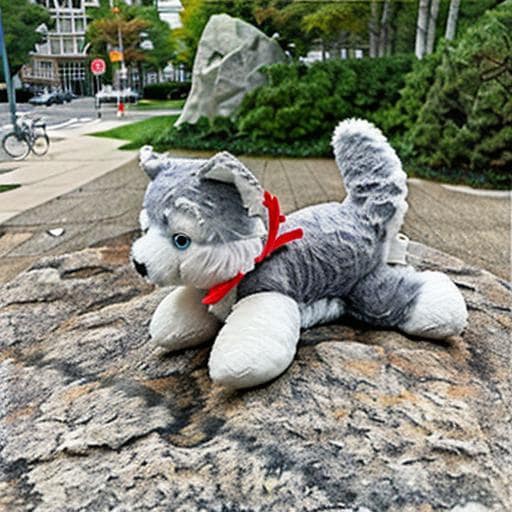} &
        \includegraphics[width=0.14\linewidth]{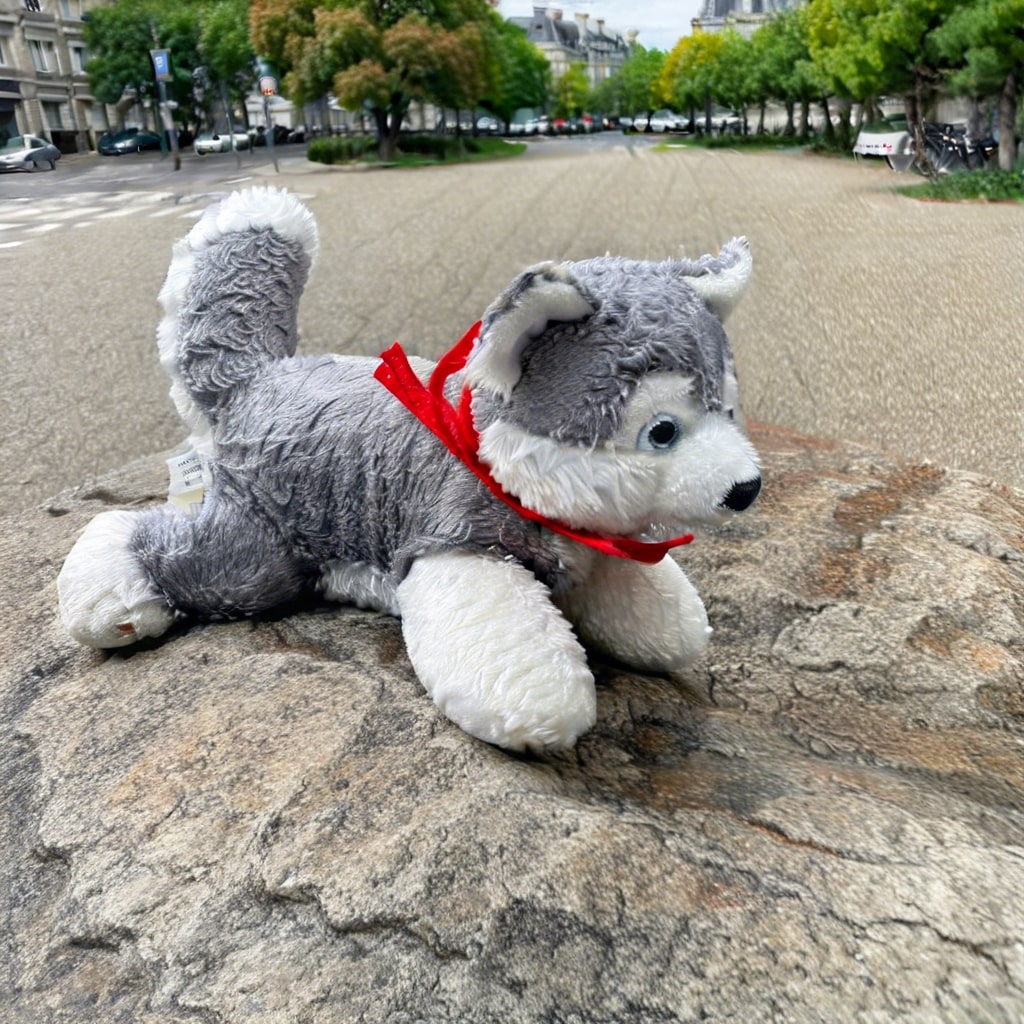} &
        \includegraphics[width=0.14\linewidth]{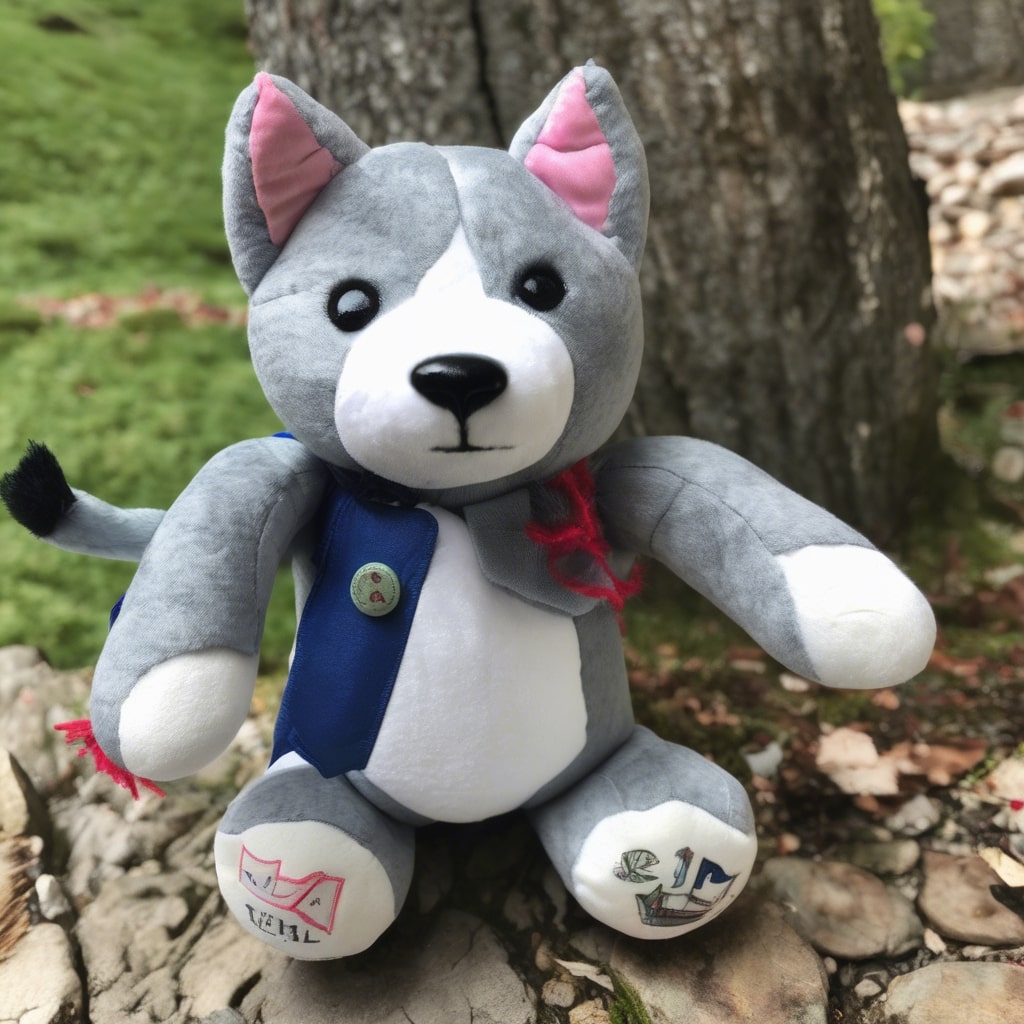} \\

        \raisebox{26pt}{\rotatebox[origin=t]{90}{\small ... on a}} &
        \raisebox{32pt}{\rotatebox[origin=t]{90}{\small wooden deck}} &
        \includegraphics[width=0.14\linewidth]{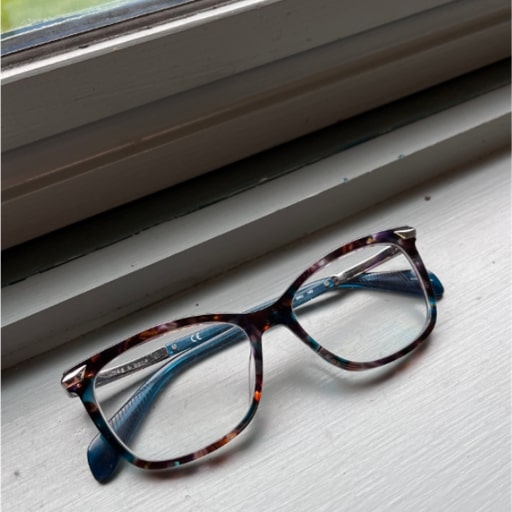} &
                \raisebox{28pt}{$\rightarrow$} &
        \includegraphics[width=0.14\linewidth]{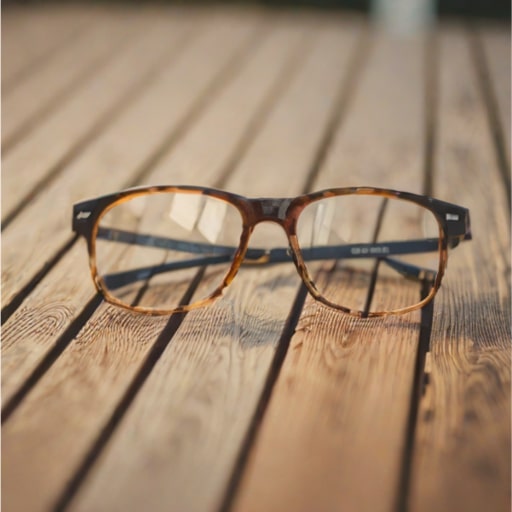} &
        \includegraphics[width=0.14\linewidth]{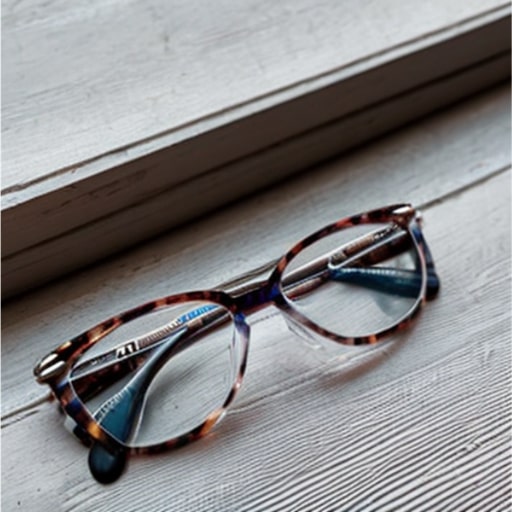} &
        \includegraphics[width=0.14\linewidth]{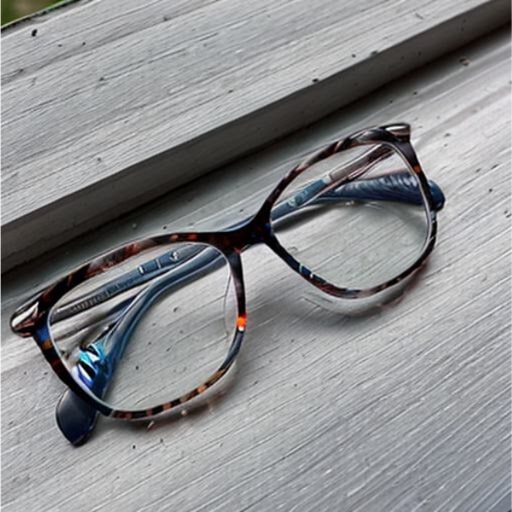} &
        \includegraphics[width=0.14\linewidth]{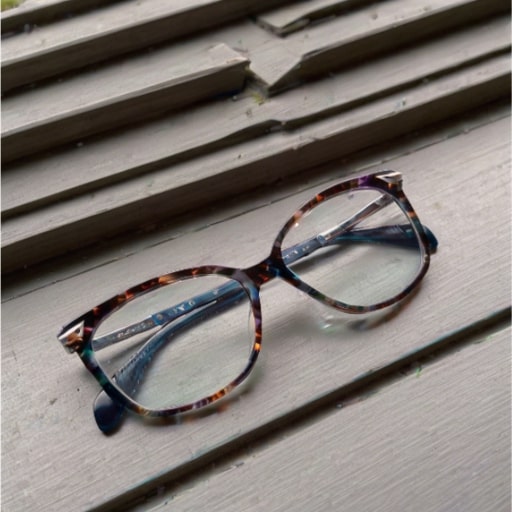} &
        \includegraphics[width=0.14\linewidth]{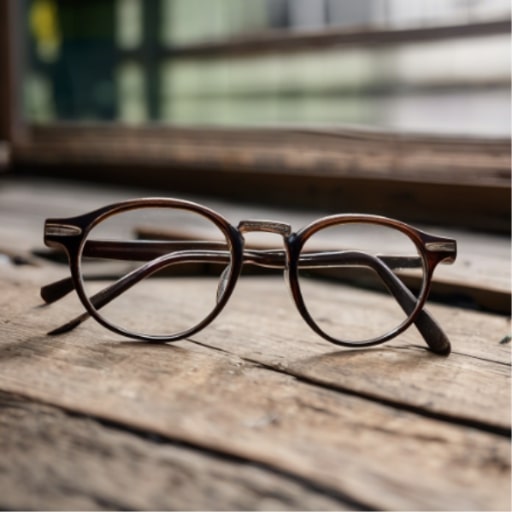} \\


        \raisebox{26pt}{\rotatebox[origin=t]{90}{\small}} &
        \raisebox{26pt}{\rotatebox[origin=t]{90}{\small ... singing karaoke}} &
        \includegraphics[width=0.14\linewidth]{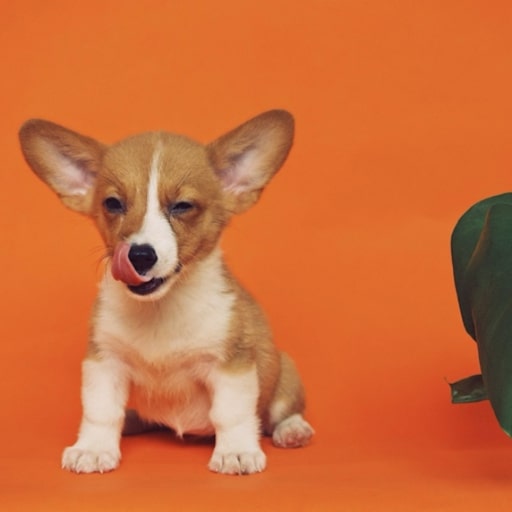} &
                \raisebox{28pt}{$\rightarrow$} &
        \includegraphics[width=0.14\linewidth]{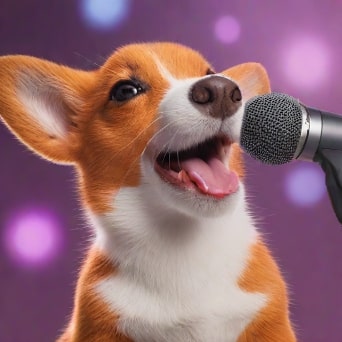} &
        \includegraphics[width=0.14\linewidth]{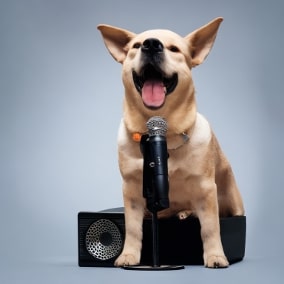} &
        \includegraphics[width=0.14\linewidth]{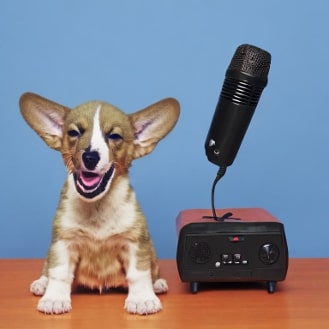} &
        \includegraphics[width=0.14\linewidth]{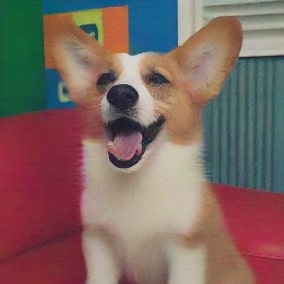} &
        \includegraphics[width=0.14\linewidth]{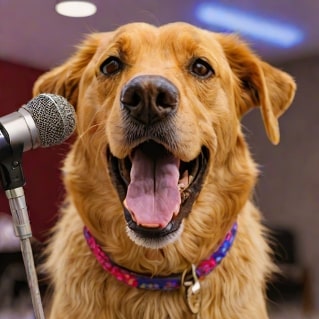} \\

    \end{tabular}
    \caption{Qualitative results for subject-driven image generation using a single subject image. The subject image is shown on the left, followed by the given prompt and the generated results from our method and various baselines.} 
    \label{fig:qual_generation}
\end{figure*}

 \begin{figure*}[t!]
    \centering
    \setlength{\tabcolsep}{1pt}
    \begin{tabular}{cccccccc }
        Input Image & Subject Image & & Ours & TIGIC & SwapAnything & DreamBooth & TI \\

        \includegraphics[width=0.12\linewidth]{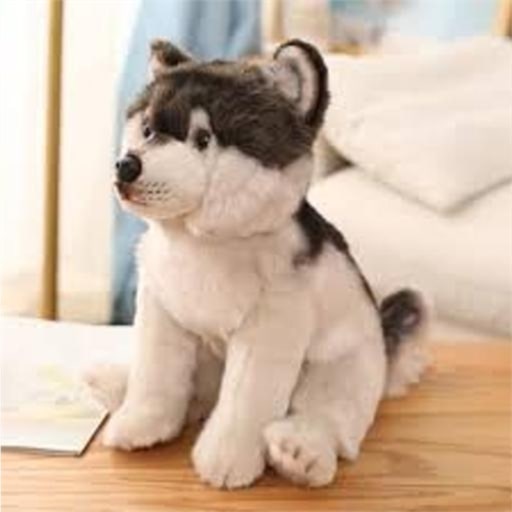} &
        \includegraphics[width=0.12\linewidth]{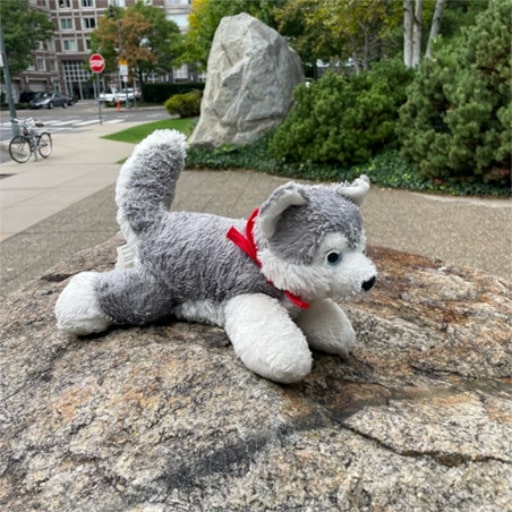} &
        \raisebox{28pt}{$\rightarrow$} &
        \includegraphics[width=0.12\linewidth]{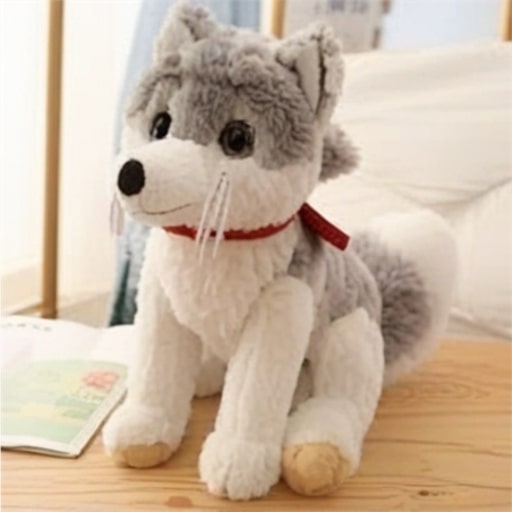} &
        \includegraphics[width=0.12\linewidth]{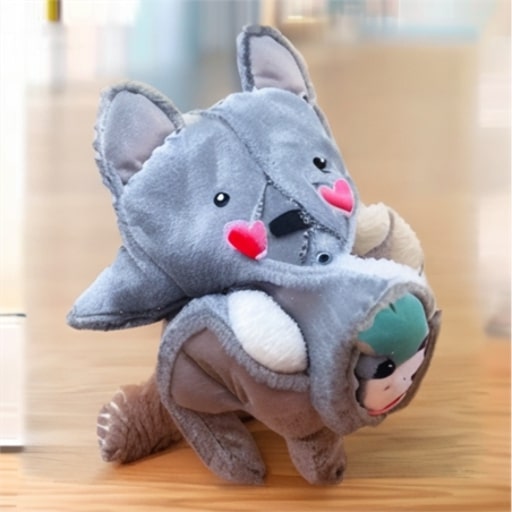} &
        \includegraphics[width=0.12\linewidth]{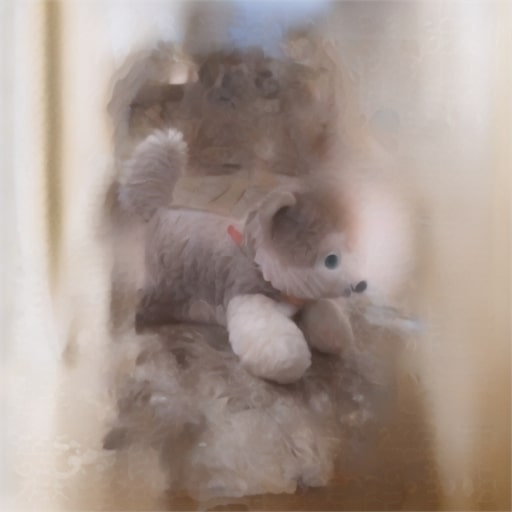} &
        \includegraphics[width=0.12\linewidth]{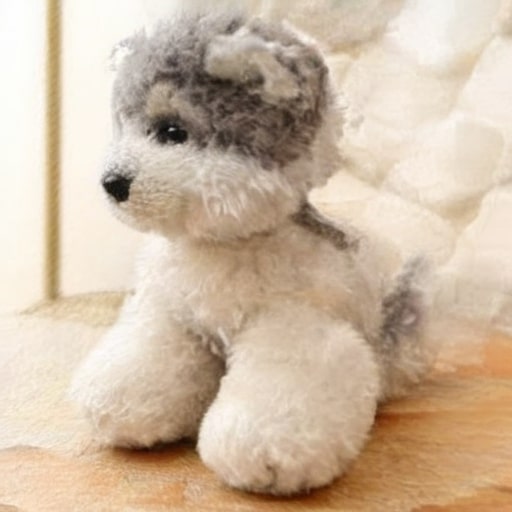} &
        \includegraphics[width=0.12\linewidth]{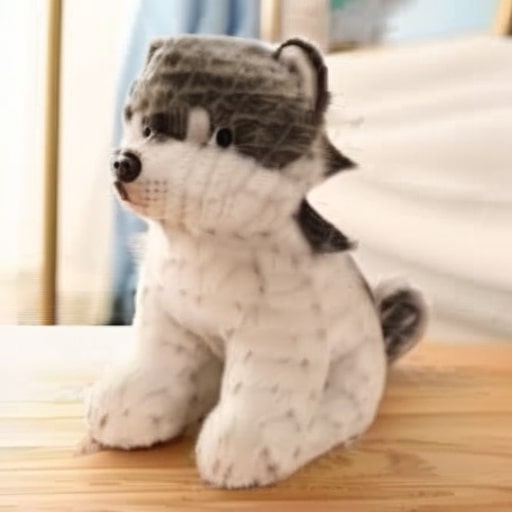} \\

        \includegraphics[width=0.12\linewidth]{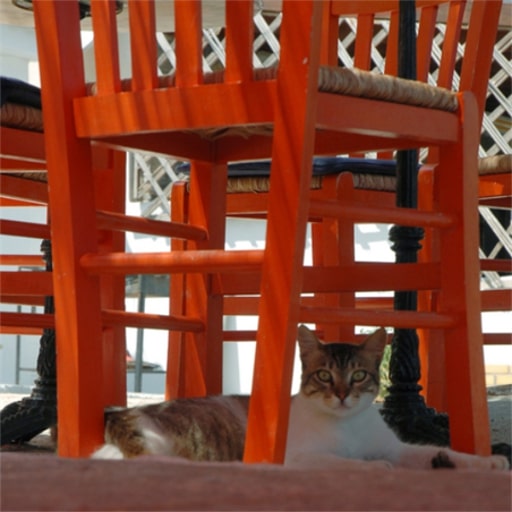} &
        \includegraphics[width=0.12\linewidth]{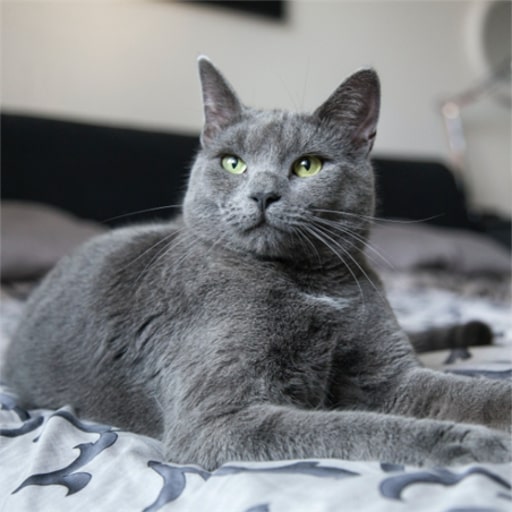} &
        \raisebox{28pt}{$\rightarrow$} &
        \includegraphics[width=0.12\linewidth]{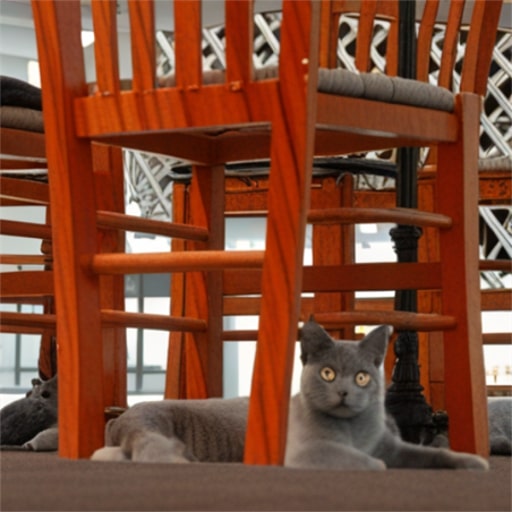} &
        \includegraphics[width=0.12\linewidth]{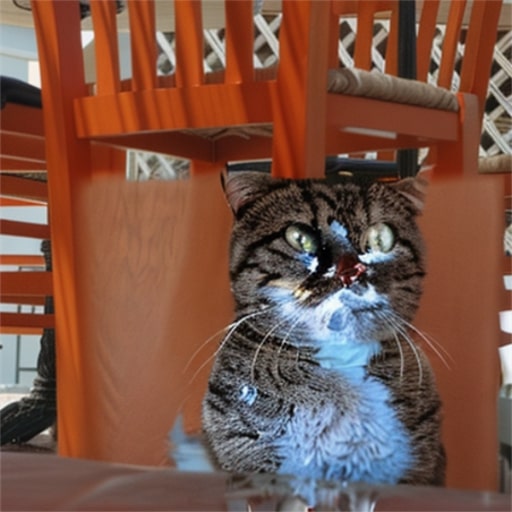} &
        \includegraphics[width=0.12\linewidth]{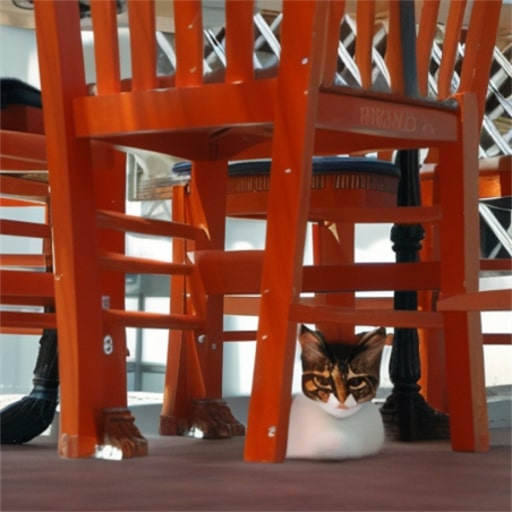} &
        \includegraphics[width=0.12\linewidth]{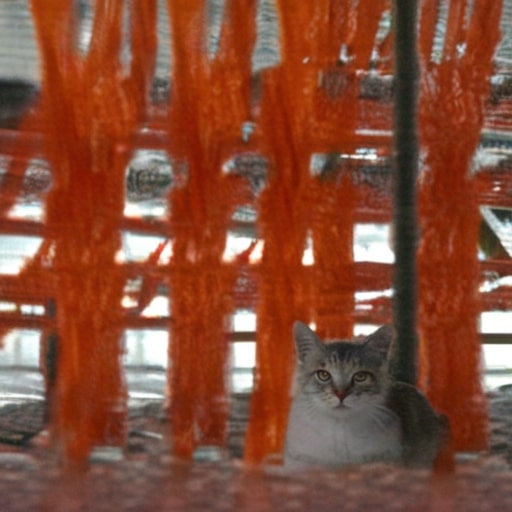} &
        \includegraphics[width=0.12\linewidth]{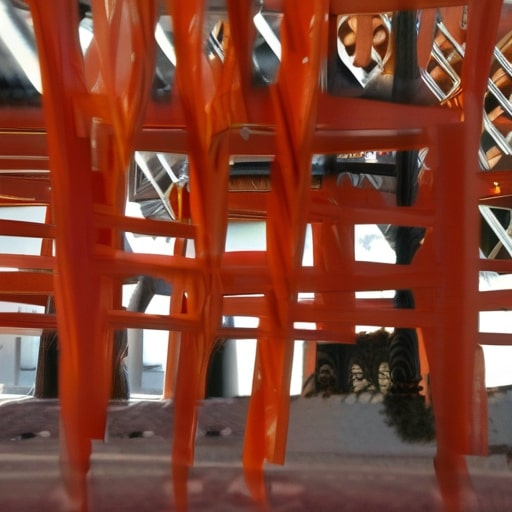} \\

        \includegraphics[width=0.12\linewidth]{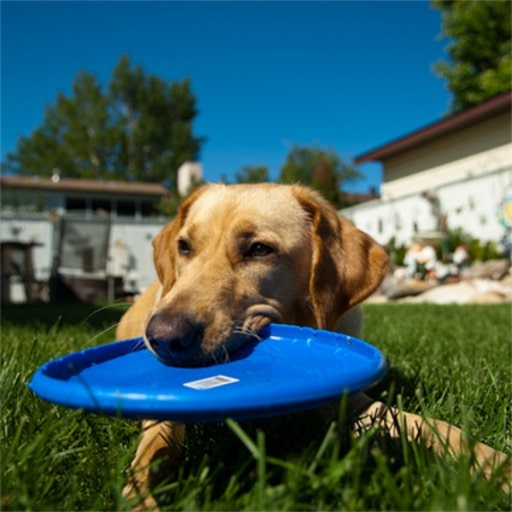} &
        \includegraphics[width=0.12\linewidth]{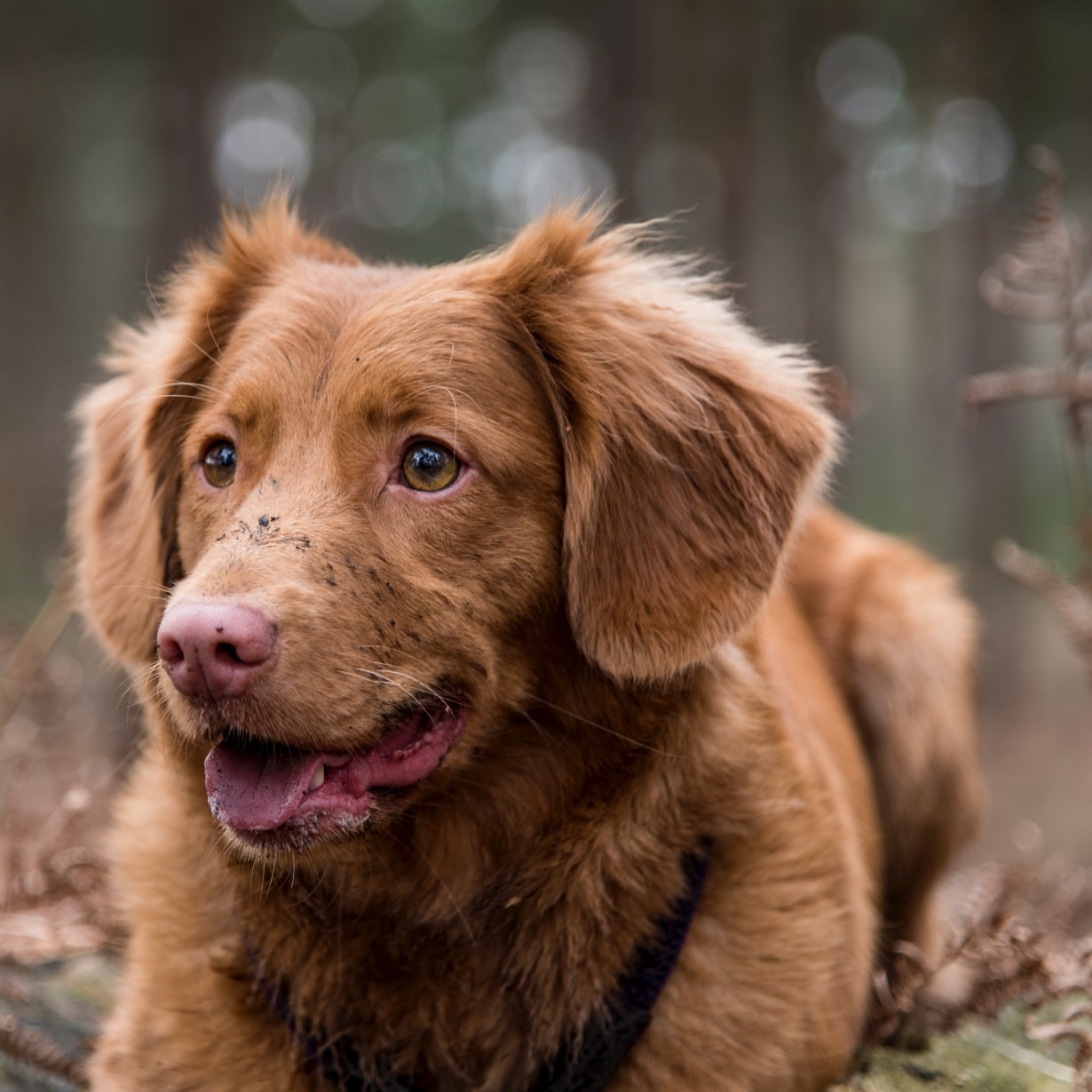} &
        \raisebox{28pt}{$\rightarrow$} &
        \includegraphics[width=0.12\linewidth]{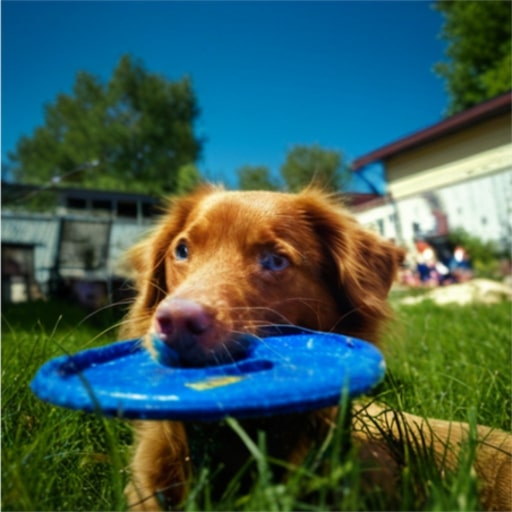} &
        \includegraphics[width=0.12\linewidth]{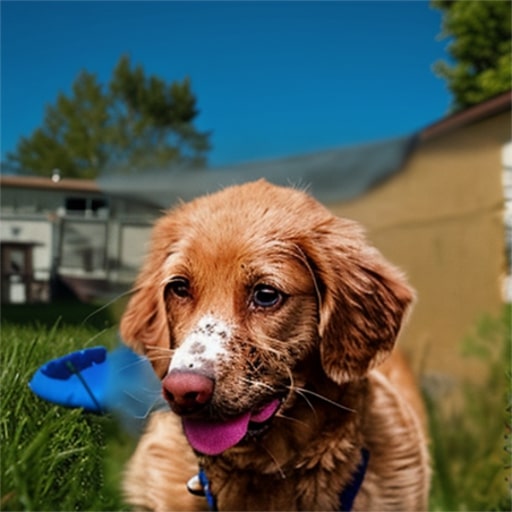} &
        \includegraphics[width=0.12\linewidth]{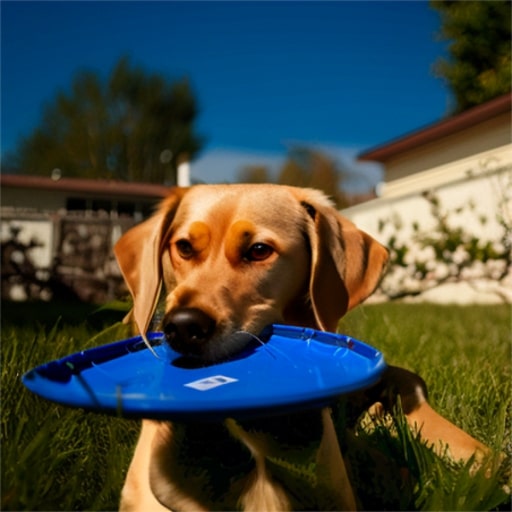} &
        \includegraphics[width=0.12\linewidth]{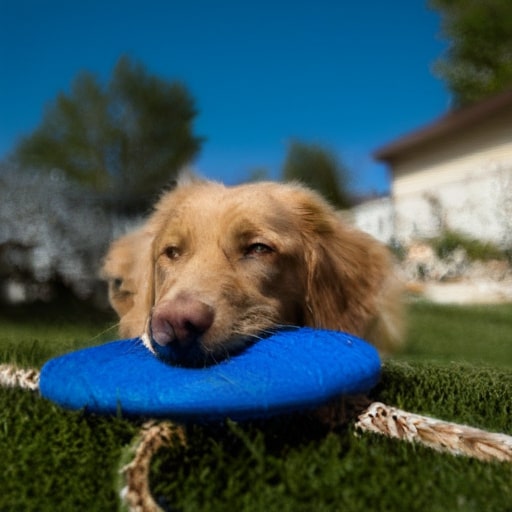} &
        \includegraphics[width=0.12\linewidth]{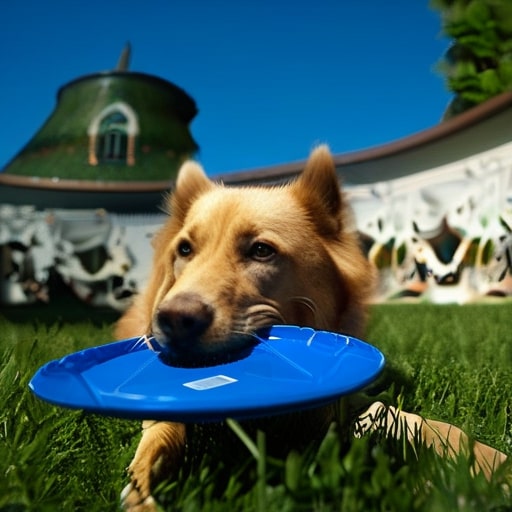} \\


    \end{tabular}
    \caption{
   Qualitative results for subject-driven image editing using a single subject image. Each row shows an original input image to be edited, a reference subject image, and results generated by our method \siso{} and four baselines.}
    \label{fig:qual_editing}
\end{figure*}

\begin{figure*}
    \centering
    \setlength{\tabcolsep}{1pt}
    \large
    \begin{tabular}{cccccc@{\hspace{10pt}}cc@{\hspace{10pt}}cc}
        & & & & \multicolumn{2}{c}{\textbf{SDXL Turbo}} & \multicolumn{2}{c}{\textbf{FLUX Schnell}} & \multicolumn{2}{c}{\textbf{Sana}} \\

        & & Subject Image & & Ours & DreamBooth & Ours & DreamBooth & Ours & DreamBooth \\

        \raisebox{26pt}{\rotatebox[origin=t]{90}{\small ... reading a book}} &
        \raisebox{26pt}{\rotatebox[origin=t]{90}{\small wearing pink glasses}} &
        \includegraphics[width=0.13\linewidth]{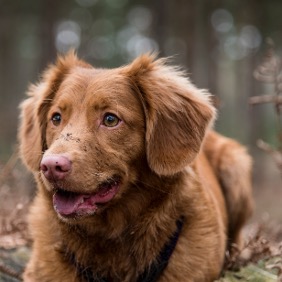} &
                \raisebox{28pt}{$\rightarrow$} &
        \includegraphics[width=0.13\linewidth]{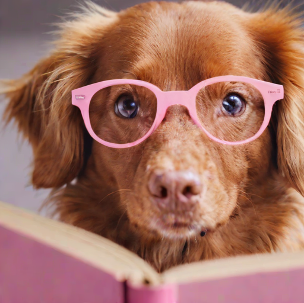} &
        \includegraphics[width=0.13\linewidth]{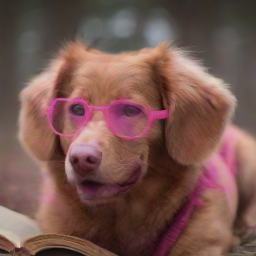} &
        \includegraphics[width=0.13\linewidth]{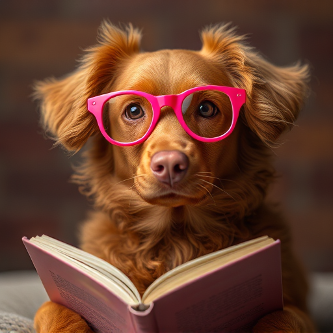} &
        \includegraphics[width=0.13\linewidth]{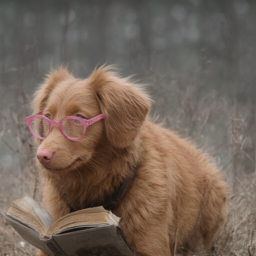} &
        \includegraphics[width=0.13\linewidth]{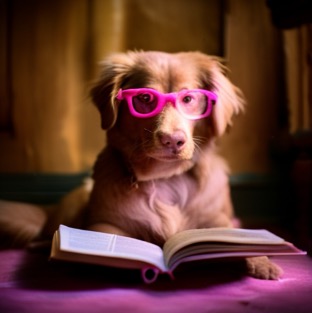} &
        \includegraphics[width=0.13\linewidth]{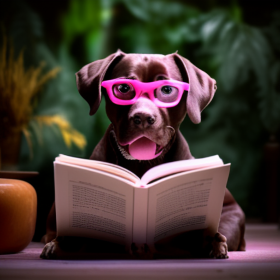} \\

        \raisebox{26pt}{\rotatebox[origin=t]{90}{\small two ... }} &
        \raisebox{26pt}{\rotatebox[origin=t]{90}{\small in a restaurant}} &
        \includegraphics[width=0.13\linewidth]{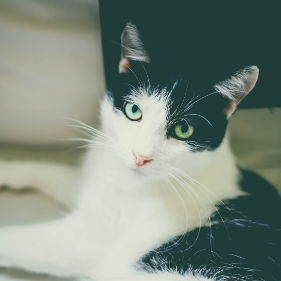} &
                \raisebox{28pt}{$\rightarrow$} &
        \includegraphics[width=0.13\linewidth]{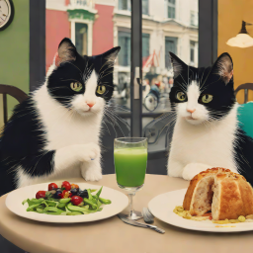} &
        \includegraphics[width=0.13\linewidth]{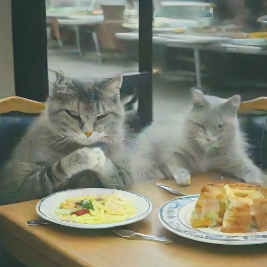} &
        \includegraphics[width=0.13\linewidth]{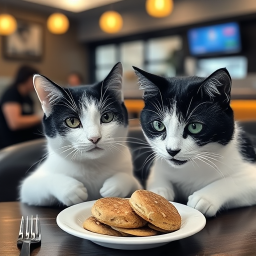} &
        \includegraphics[width=0.13\linewidth]{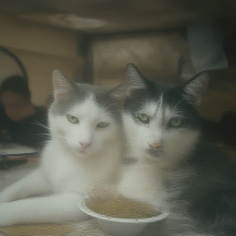} &
        \includegraphics[width=0.13\linewidth]{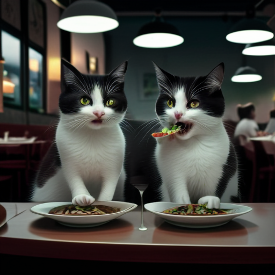} &
        \includegraphics[width=0.13\linewidth]{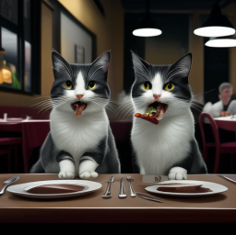} \\

    \end{tabular}
    \caption{Subject-driven image generation using three backbone models (single reference image)} 
    \label{fig:multi_gen_models}
\end{figure*}

\paragraph{Prompt adherence.}
In image generation, we also measure alignment with the input prompt using CLIP-T, %
the CLIP score between the generated image and the input prompt.

\paragraph{Diversity.}  
Single-image concept learning often leads to overfitting, limiting diversity in generated images due to reconstruction loss. To quantify this, we compute the mean squared error (MSE) between generated and subject images.  

\paragraph{Background Preservation.}  
For image editing, maintaining the background while altering the subject is crucial. We assess this using LPIPS~\cite{zhang2018unreasonable}, where lower scores indicate higher similarity. To exclude the edited region, we mask the subject using Grounding DINO and SAM~\cite{kirillov2023segment} before computing LPIPS.

\subsection{Quantitative Results}

\paragraph{Image Generation.}
Table~\ref{tab:generation_small} shows results for image generation, comparing \siso{} to two subject-driven baselines that typically learn from multiple subject images but are tested here with a single reference. \siso{} significantly improves naturalness metrics, suggesting that baselines degrade image quality due to overfitting. Additionally, \siso{} enhances prompt adherence while maintaining subject identity. This suggests that aligning the image directly, rather than splitting the process into separate optimization and generation stages, improves identity preservation—albeit with a slight trade-off in naturalness or prompt accuracy.

Next, in Table~\ref{tab:generation}, we further evaluate the adaptability of different models for subject-driven generation using a single image. To our knowledge, DreamBooth is the only baseline that can be easily adapted across models, as others are tailored specifically for Stable Diffusion 2.1. Our results show that our method outperforms DreamBooth in identity preservation for FLUX and Sana. Although DreamBooth achieves better identity preservation with SDXL-Turbo, this is mainly due to overfitting, as indicated by the diversity metrics (0.05 vs. 0.11).

\paragraph{Image Editing.}
Table~\ref{tab:edit} compares our approach against subject-driven image editing baselines. TIGIC blends the subject into the image during diffusion, often resulting in background corruption (0.22 vs. 0.14). SwapAnything learns the subject concept, but when only a single subject image is used, its identity preservation significantly declines (0.55 vs. 0.80 on DINO). Additionally, naturalness metrics are low, with an FID score of 185.7, suggesting that fewer input subject images can substantially drop image quality. 

\paragraph{Ablation.}

In Table~\ref{tab:ablation-gen}, we examine prompt simplification. We observe a trade-off: Simplifying the prompt improves adherence, while direct optimization with the full prompt better preserves subject identity.

Table~\ref{tab:ablation-edit} evaluates the impact of background preservation loss (Eq.~\ref{eq:bg_loss}) on editing. Adding this loss improves background consistency (LPIPS: 0.14 vs. 0.18) without compromising identity preservation. We also assess using DINO and IR in an ensemble, which enhances identity preservation with only slightly reduced background consistency (LPIPS: 0.12 vs. 0.14).

\noindent\newline\textbf{User study.}
In addition to automated metrics, we conducted a user study to measure identity preservation, background preservation, prompt adherence, and naturalness. We used Amazon MTurk for 100 images, with five raters per image. See full details in the appendix (Sec. \ref{sec:user}). Two user studies were conducted separately, one for editing and one for generation, comparing our method against the best available baseline of each task.  

The results of the user study are given in Table~\ref{tab:human}. 
For editing, TIGIC better preserves subject identity because it often acts almost as a copy-paste of the subject into the given input image. This is reflected in \siso{} obtaining higher scores for both naturalness and background preservation, with win rates of 58\% and 60\%, respectively. In the generation task, we see a slight improvement for the baseline in subject preservation (47\% win rate). However, \siso{} produces significantly more natural images (65\% win rate) and shows prompt adherence (69\% win rate).

\subsection{Qualitative Results}

We begin by showing the results of our generative model compared to popular baselines (Fig.~\ref{fig:qual_generation}). We evaluate subject-driven image generation on three subjects: a plush toy, glasses and a dog. Only our method correctly places the plush in Paris, while others overfit to the input image. Textual Inversion (TI) avoids this but fails to capture identity. Similar issues arise with the glasses, where most methods retain background elements, except ours and TI, though TI lacks detail. Our method preserves subject identity while generating diverse backgrounds. In the final row, the baselines fail to depict the subject and follow the prompt.

In Fig.~\ref{fig:qual_editing}, we compare baselines for image editing by learning subject concepts, inverting, and regenerating images. In the first row, our method accurately preserves the wolf plush, while baselines either blend unnaturally (TIGIC), leak background details (SwapAnything), or distort both subject and background (DreamBooth, TI). In the second row, our method correctly replaces the black cat, though with a slight eye color mismatch, while baselines fail entirely. In the third row, all methods perform better, but ours best preserves the background.

In Fig.~\ref{fig:multi_gen_models}, we present subject-driven image generation using a single reference image with SDXL Turbo, FLUX Schnell, and Sana models. DreamBooth, the only baseline adaptable across models, shows several limitations when trained on a single image: (i) low diversity, with generations closely resembling the subject (e.g., the dog generated by SDXL and FLUX, the cat by FLUX), (ii) artifacts and unnatural attributes (e.g., the cats generated by SDXL and FLUX), and (iii) poor identity preservation (e.g., the dog in Sana).

We also assess the stability of our method using various seeds (see Figures ~\ref{fig:seed_stability_gen} and ~\ref{fig:seed_stability_edit} in the appendix).

\section{Conclusion}

We present \siso{}, a novel optimization technique that employs a single subject image and enables subject-driven image generation and subject-driven image editing by leveraging pre-trained image similarity score models. We show that in all previous baselines, enabling such capability with a single image in an existing diffusion model is far from being solved. While our method still has room for improvement in subject identity preservation, it opens up a new research thread that may make the personalization of image generators as simple as possible with the use of only a single image.

\section{Acknowledgments}
This work was supported by a Vatat datascience grant.

\bibliography{bibliography}

\clearpage
\appendix

In this supplementary material, we present additional experiment results. The supplementary comprises the following subsections:

\begin{enumerate}
    \item Sec.~\ref{sec:diffusion_inversion}, details the inversion method we used for image editing.
    \item Sec.~\ref{sec:user}, details about the user study.
    \item Sec.~\ref{sec:early_stopping}, details about early-stopping method used in our experiments.
    \item Sec.~\ref{sec:baselines}, details about the implementation of the baselines.
    \item Sec.~\ref{sec:adaptation}, details about adaptation of \siso\ to various bacbone models.
    \item Sec.~\ref{sec:faces}, details about the attempt to use \siso\ for subject-driven face swapping.
\end{enumerate}

\section{Diffusion Inversion}
\label{sec:diffusion_inversion}
We employ ReNoise for diffusion inversion in our image editing solution. ReNoise hyperparameters include strength, calibrating noise addition, balancing reconstruction, and editability. High values harm reconstruction while improving the ability to edit, and low values hinder object changes but improve reconstruction. We tuned the default setting from 1 to 0.75 in all experiments. Although this setting slightly reduces editing potential, subject-driven editing demands changes to the subject, not the background. Thus, this value empirically proved optimal for both reconstruction and subject editing without altering the background.

\section{User Study}
\label{sec:user}
\begin{figure}[ht]
    \centering
    \includegraphics[width=\columnwidth]{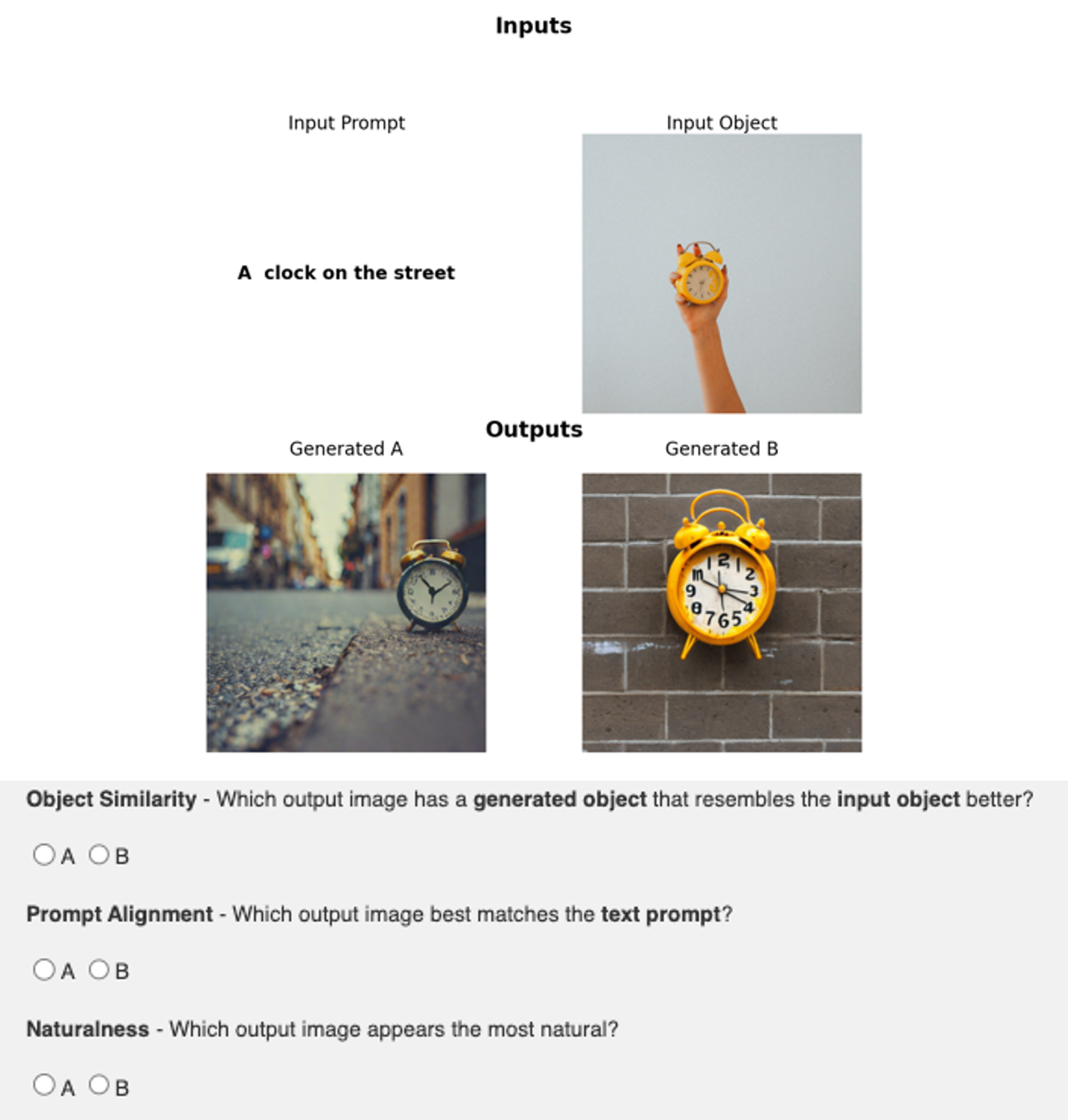}
    \caption{Illustration of the user study interface for Subject-driven image generation task.}
    \label{fig:user-study}
\end{figure}

According to the task, workers in Amazon MTurk were presented with a subject image, a condition (a prompt or an input image), and two generated images - one from SISO and the other from the baseline.  The study was conducted on 100 images from the benchmark, with five workers rated each image. The method used for the study was Two-alternative forced choice, where raters must choose the preferred output between two options. In our case, the workers were presented three questions per image. Each question requested the worker to choose between two generated images (the order between the generated images was randomly picked). For subject-driven image generation, the questions tested the following criteria: (i) object similarity (what we refer in the paper as identity preservation), (ii) prompt alignment (what we refer as prompt adherence) and (iii) naturalness. See Fig. \ref{fig:user-study} for illustration of the user study interface.

\section{Early Stopping}
\label{sec:early_stopping}
\siso\ generates a well-formed image at each iteration, rather than noisy latent. This enables using the method in an interactive manner. One option is to display images from all iterations and stop the optimization process when a satisfactory result is obtained. To achieve a fully automated process, we used a simple early-stopping strategy, where the process ends if the loss has not improved by $x$ percent on the last $n$ iterations. Specifically, we set $x = 3$ and $n = 7$ in all of our experiments, both for generation and editing.

\section{Baselines}
\label{sec:baselines}
Here, we describe how we implemented the baselines used in the paper. 

\paragraph{Subject-driven image generation.} We compared our method against three baselines: (i) DreamBooth, which fine-tunes the diffusion model parameters according to a set of reference images.  We used the code given in Diffusers ~\cite{von-platen-etal-2022-diffusers} library for all different base models (SDXL, FLUX, and Sana). (ii) AttnDreamBooth, which improves on DreamBooth with a three-stage process, optimizing a textual embedding, cross-attention layers, and the U-Net. (iii) ClassDiffusion, which utilizes a semantic preservation loss. For both AttnDreamBooth and ClassDiffusion we used the official implementation published by the authors, using their default hyper-parameters.

\paragraph{Subject-driven image editing.} We compared our method with two baselines: (i) SwapAnything, which employs masked latent blending and appearance adaptation.  (ii) TIGIC, a training-free technique that uses an attention-blending strategy during the denoising process. TIGIC was initially designed for a subject insertion, where the user wants to insert the subject to an empty area in the input image. To adapt to the subject replacement task, we used a state-of-the-art inpainting model (LaMa\footnote{\url{https://github.com/advimman/lama}}) to remove the original object and then applied TIGIC. For both methods, we used the official implementation published by the authors, using their default hyper-parameters.

\section{Adaptation to Various Backbone Models}
\label{sec:adaptation}
\begin{figure}
    \centering
    \setlength{\tabcolsep}{1pt}
    \scriptsize{
    \small
    \begin{tabular}{cccc}
        Subject Image & & Output 1 & Output 2 \\

        \includegraphics[width=0.27\linewidth]{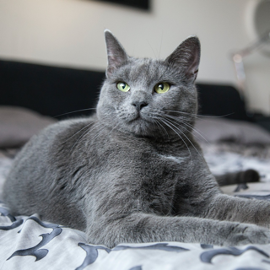} &
        \raisebox{22pt}{$\rightarrow$} &
        \includegraphics[width=0.27\linewidth]{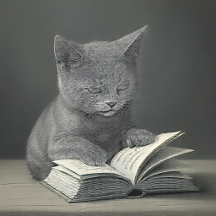} &
        \includegraphics[width=0.27\linewidth]{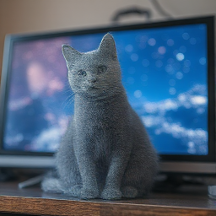} \\

        \includegraphics[width=0.27\linewidth]{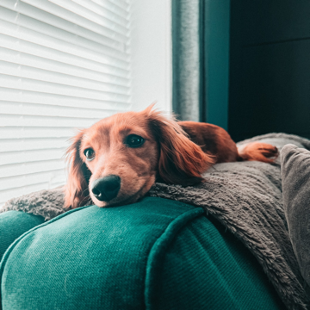} &
        \raisebox{22pt}{$\rightarrow$} &
        \includegraphics[width=0.27\linewidth]{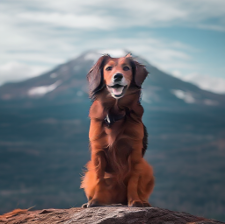} &
        \includegraphics[width=0.27\linewidth]{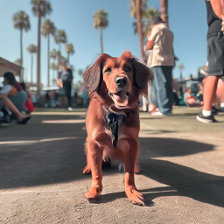} \\
        
    \end{tabular}
    }
    \caption{
        Optimizing on FLUX schnell using four denoising steps results in low quality images.
    }
    \label{fig:flux_low}
\end{figure}

A key advantage of \siso\ is its ability to be used with different backbone models with limited adaptation. In this section, we will describe the main differences in implementation between the different backbones we used (SDXL-Turbo, FLUX schnell, Sana). First, SDXL-Turbo and FLUX schnell are distilled versions, meaning that they generate images using a small number of steps (1-4). Sana, on the other hand, does not have a distilled version and requires ~20 steps to generate a high-quality image. We found that when using distilled versions, backpropogating through the final denoising step is sufficient. However, when using a non-distilled version, like Sana, it may be beneficial to backpropagate through more than one denoising step. Specifically, we set the number of steps to backpropogate through to 3. 
Also, even when using a distilled version, the number of denoising steps used in each iteration may be important, and different models behave differently in this context. We will denote this number as $t$. SDXL-Turbo is less noisy to different values of $t$, but FLUX schnell showed a significant difference when using various values of $t$. More specifically, setting $t >= 2$ resulted in low-quality generated images, even when trying to backpropogate through more denoising steps (see Fig.~\ref{fig:flux_low}). However, FLUX schnell generates blurred images when used with one denoising step. A naive approach to overcome the blurriness is to use a model trained for up-scaling resolution. But this requires loading another model and may complicate the process. We solved the issue using the training simplification (Sec.~\ref{sec:generation} in the paper). Although the weights were optimized using $t=1$, they can be used in inference with different values of $t$, thus producing high-quality images. 

\section{Subject-driven Face Swapping}
\label{sec:faces}
\begin{figure}[h]
    \centering
    \setlength{\tabcolsep}{1pt}
    \scriptsize{
    \small
    \begin{tabular}{cccc }
        Subject Image & & Input Image & Output \\

        \includegraphics[width=0.27\linewidth]{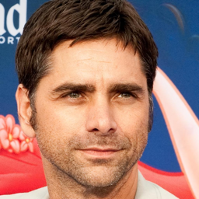} &
        \raisebox{26pt}{$\rightarrow$} &
        \includegraphics[width=0.27\linewidth]{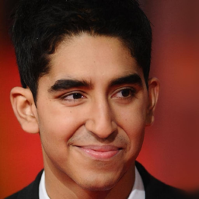} &
        \includegraphics[width=0.27\linewidth]{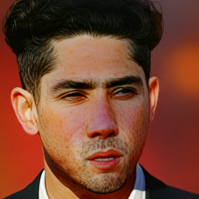} \\

        \includegraphics[width=0.27\linewidth]{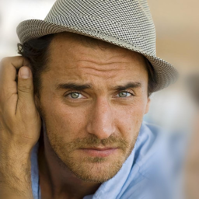} &
        \raisebox{26pt}{$\rightarrow$} &
        \includegraphics[width=0.27\linewidth]{images/faces/face_1_ref.png} &
        \includegraphics[width=0.27\linewidth]{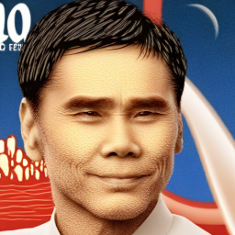} \\
        
    \end{tabular}
    }
    \caption{
        Results for subject-driven face swapping.
    }
    \label{fig:faces}
\end{figure}

A natural use-case for \siso\ is subject-driven face swapping. We tried to adapt our method to this task by using a different feature extractor more suitable for face recognition. Specifically, we employed InceptionResnet~\cite{szegedy2015going}, using the implementation from pytorch-facenet library\footnote{\url{https://github.com/timesler/facenet-pytorch}}). While this direction has a potential, it did not show satisfactory results (see Fig.~\ref{fig:faces}). 

\begin{figure*}
    \centering
    \setlength{\tabcolsep}{1pt}
    \small 
    \begin{tabular}{ccccc }
        \text{\large Subject Image} & & & & \\

        \includegraphics[width=0.16\linewidth]{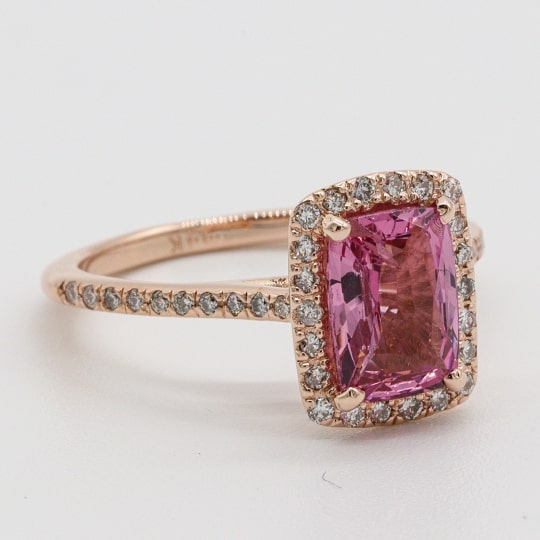} &
        \includegraphics[width=0.16\linewidth]{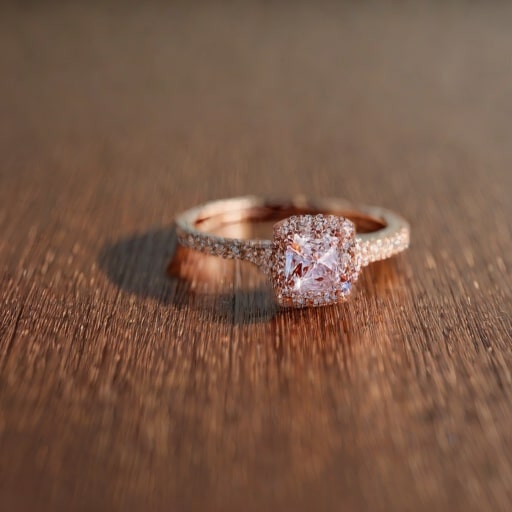} &
        \includegraphics[width=0.16\linewidth]{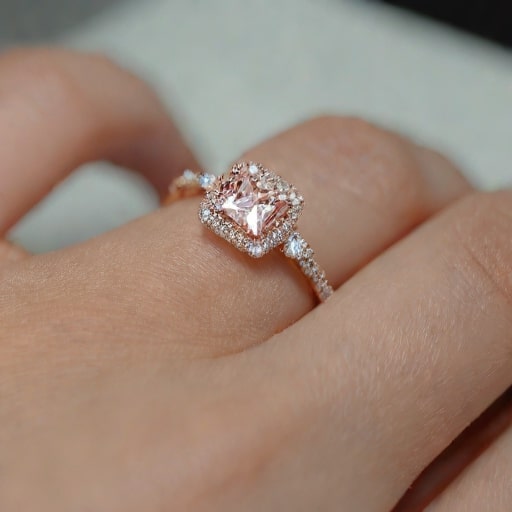} &
        \includegraphics[width=0.16\linewidth]{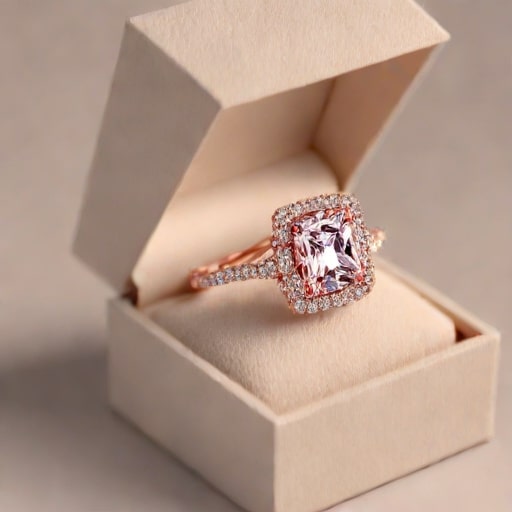} &
        \includegraphics[width=0.16\linewidth]{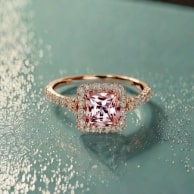} \\
        & ... on a wooden counter & ... on a finger & ... in a box & .. on a glass table \\

        \includegraphics[width=0.16\linewidth]{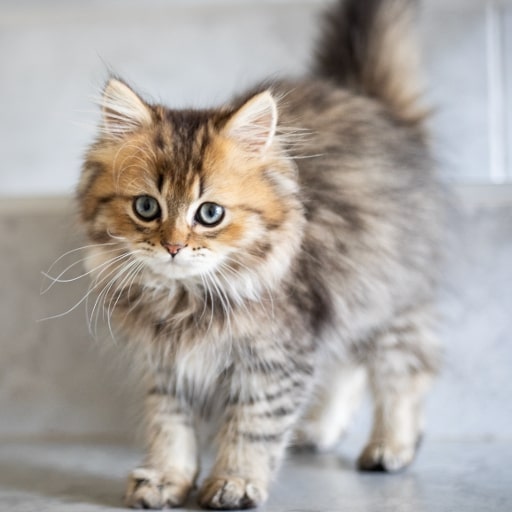} &
        \includegraphics[width=0.16\linewidth]{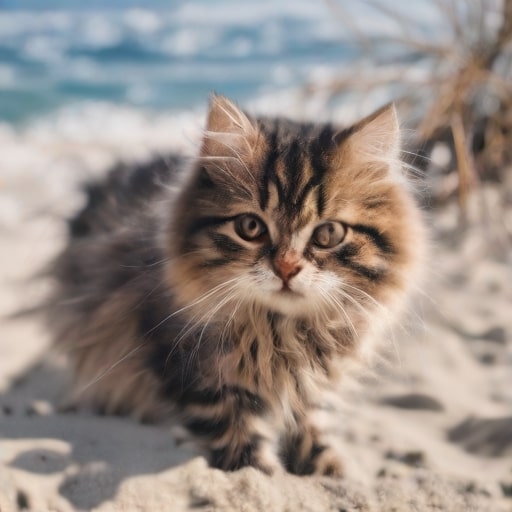} &
        \includegraphics[width=0.16\linewidth]{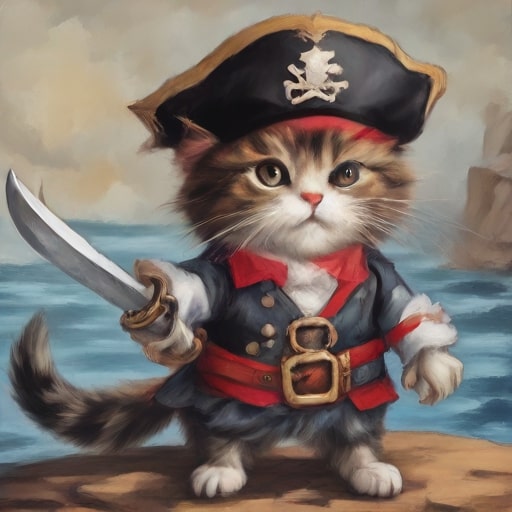} &
        \includegraphics[width=0.16\linewidth]{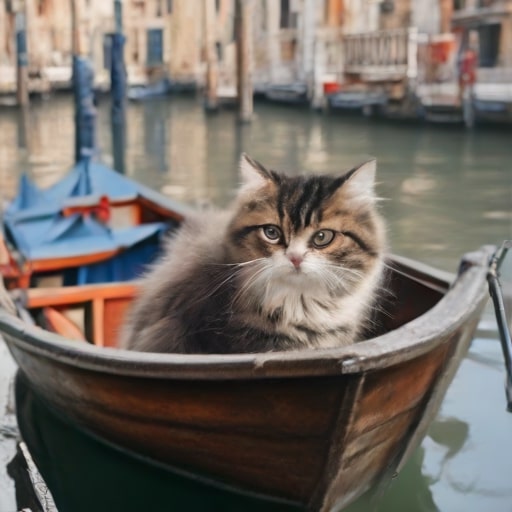} &
        \includegraphics[width=0.16\linewidth]{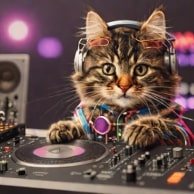} \\
        & ... in the beach & ... as a pirate & ... in a gondolla & ... as a dj \\

        \includegraphics[width=0.16\linewidth]{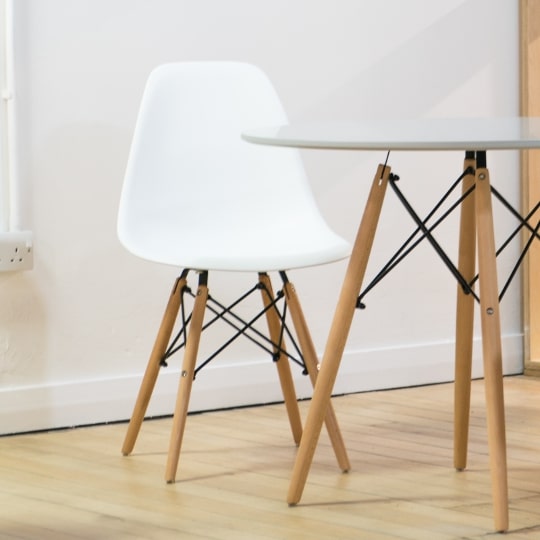} &
        \includegraphics[width=0.16\linewidth]{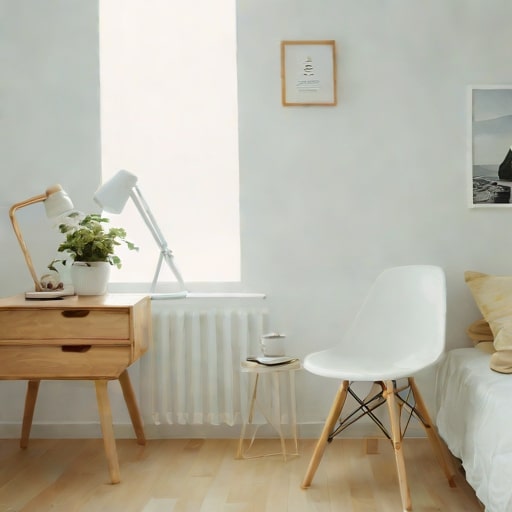} &
        \includegraphics[width=0.16\linewidth]{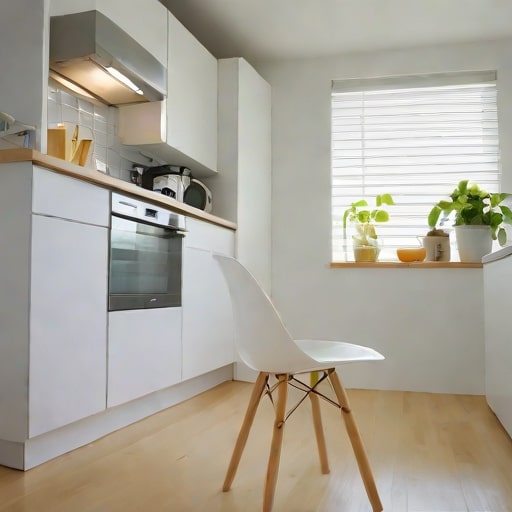} &
        \includegraphics[width=0.16\linewidth]{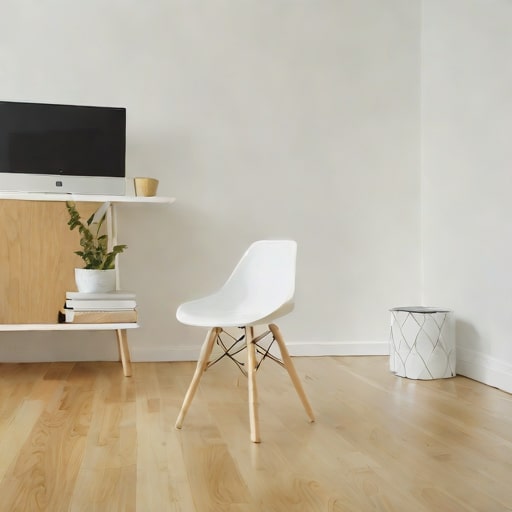} &
        \includegraphics[width=0.16\linewidth]{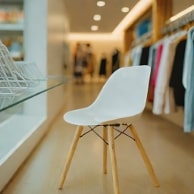} \\
        & ... in the bedroom & ... in the kitchen & ... in the living room & ... in a store \\

        \includegraphics[width=0.16\linewidth]{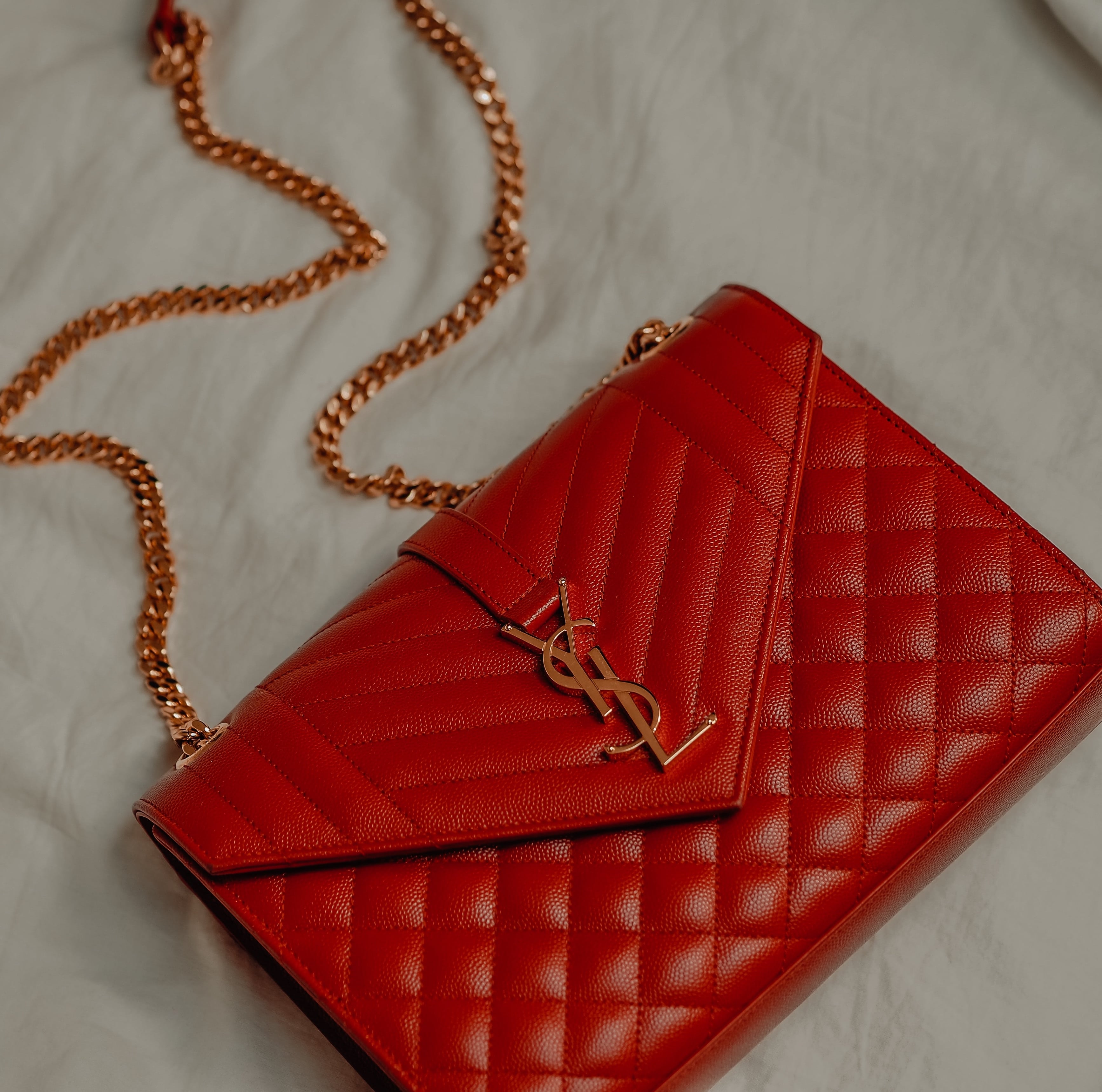} &
        \includegraphics[width=0.16\linewidth]{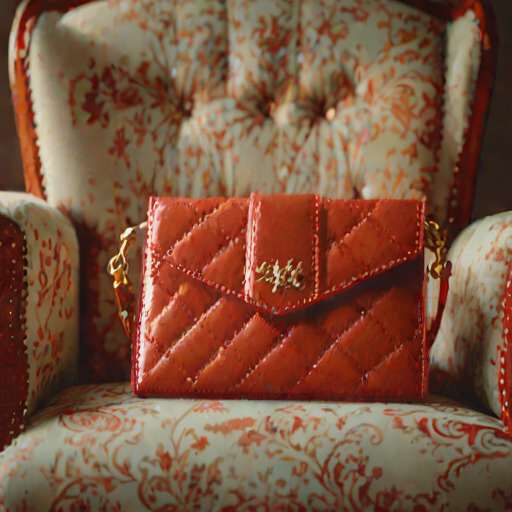} &
        \includegraphics[width=0.16\linewidth]{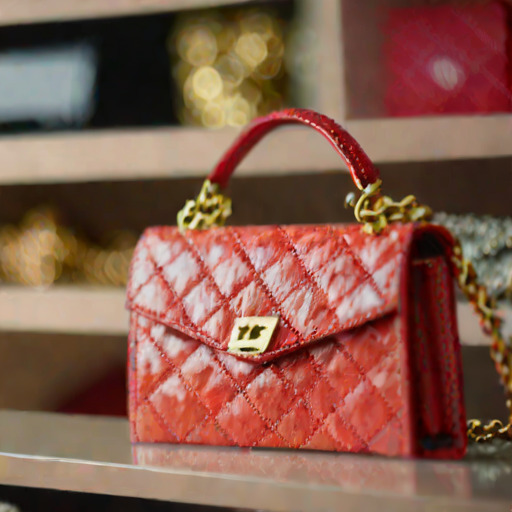} &
        \includegraphics[width=0.16\linewidth]{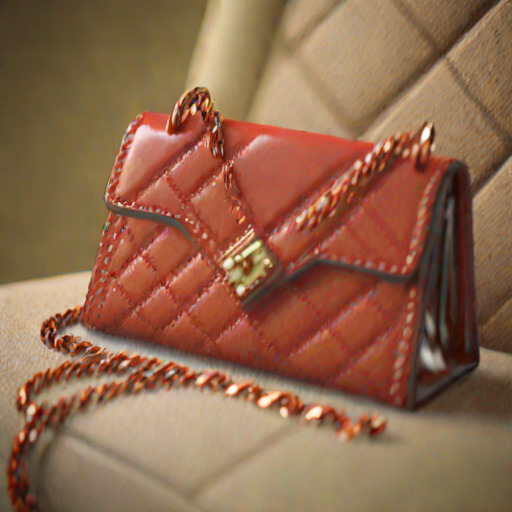} &
        \includegraphics[width=0.16\linewidth]{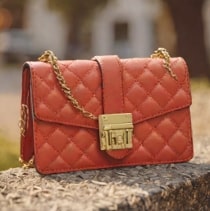} \\
        & ... on an armchair & ... on a shelf & ... on a couch & ... on a fence \\

        \includegraphics[width=0.16\linewidth]{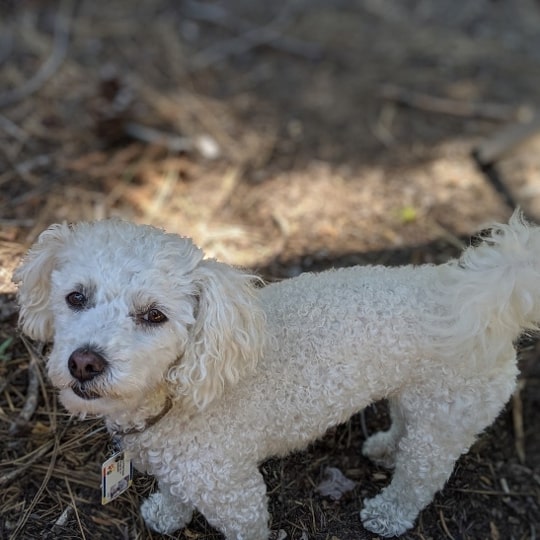} &
        \includegraphics[width=0.16\linewidth]{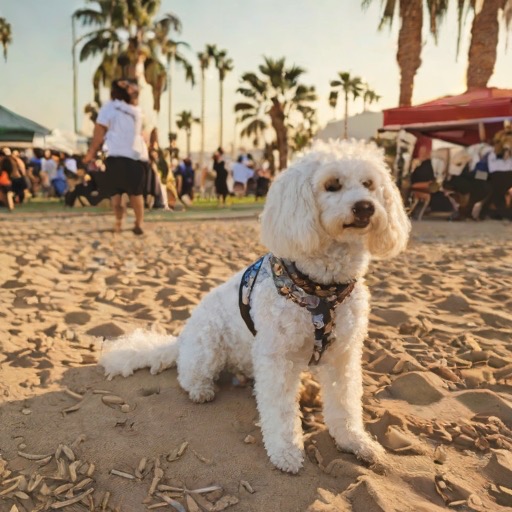} &
        \includegraphics[width=0.16\linewidth]{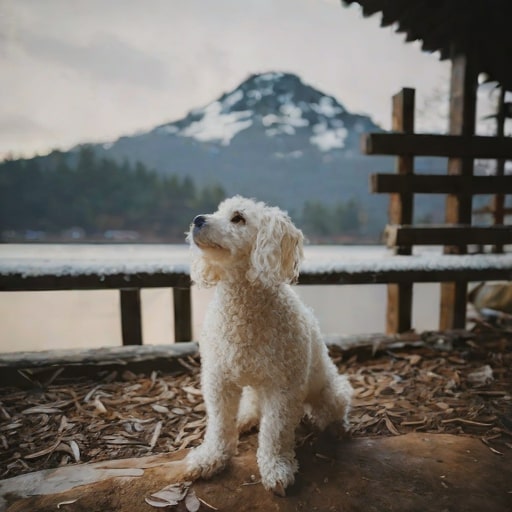} &
        \includegraphics[width=0.16\linewidth]{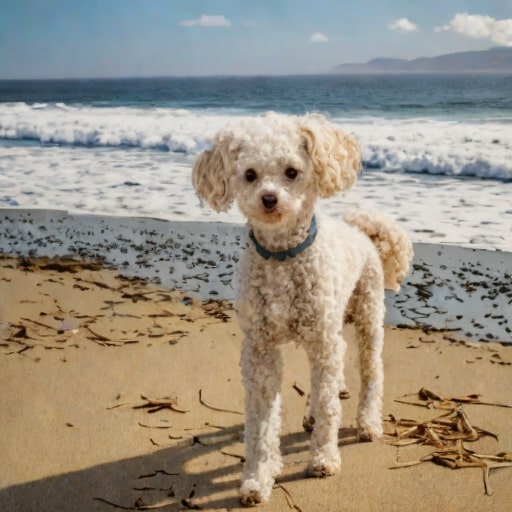} &
        \includegraphics[width=0.16\linewidth]{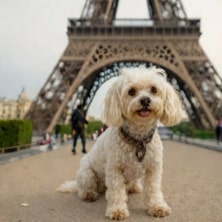} \\
        & ... in Coachella & ... in Fuji mountain & ... in the beach & ... in Paris \\

        \includegraphics[width=0.16\linewidth]{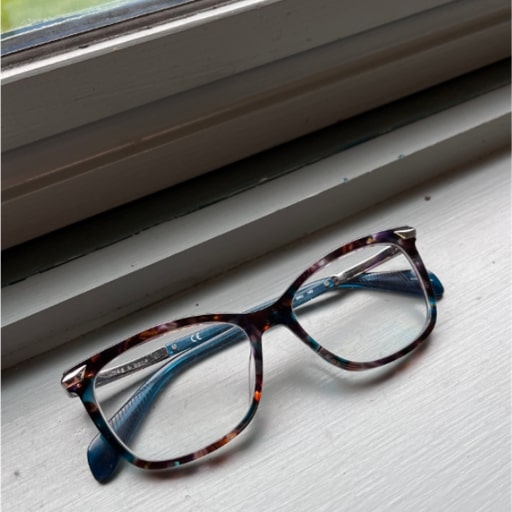} &
        \includegraphics[width=0.16\linewidth]{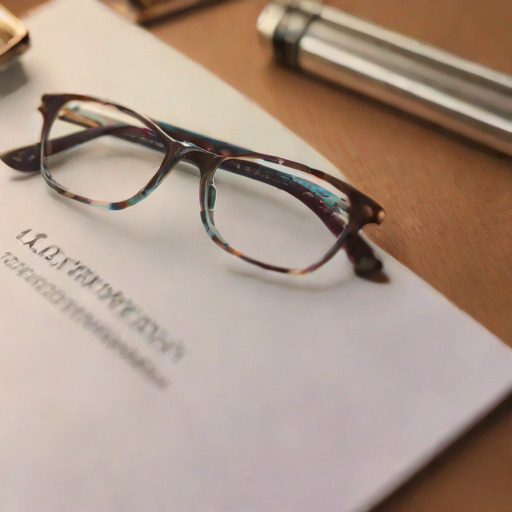} &
        \includegraphics[width=0.16\linewidth]{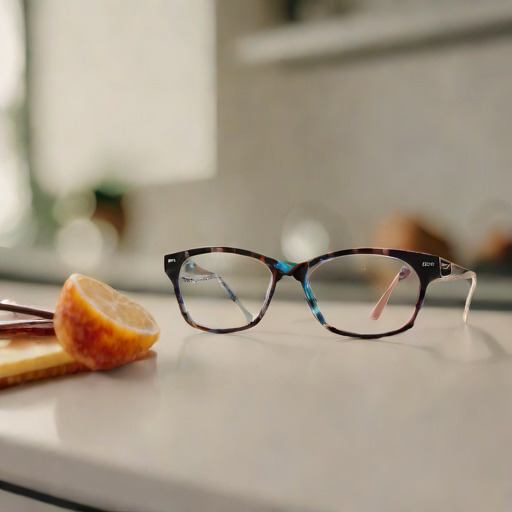} &
        \includegraphics[width=0.16\linewidth]{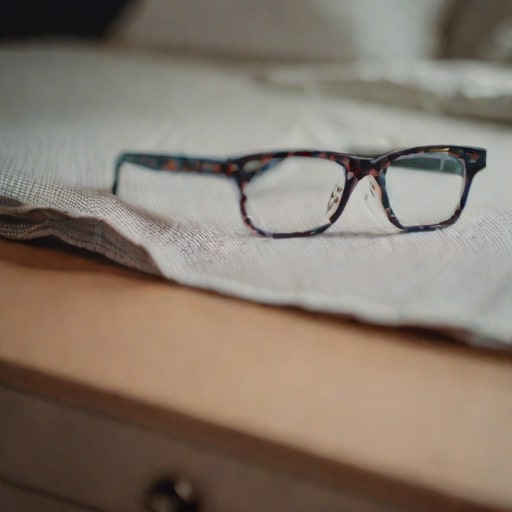} &
        \includegraphics[width=0.16\linewidth]{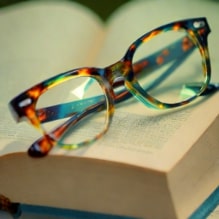} \\
        & ... on a desk & ... in the kitchen & ... on a nightstand & ... on a book \\

    \end{tabular}
    \caption{
        More Qualitative results on Subject Driven Image Generation
    }
    \label{fig:more_generation}
\end{figure*}

\begin{figure*}[t!]
    \centering
    \setlength{\tabcolsep}{1pt}
    \scriptsize{
    \large
    \begin{tabular}{cccccc }
        Subject Image & & Input Image & Output & Input Image & Output \\

        \includegraphics[width=0.17\linewidth]{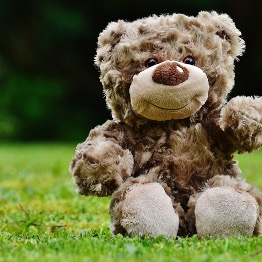} &
        \raisebox{36pt}{$\rightarrow$} &
        \includegraphics[width=0.17\linewidth]{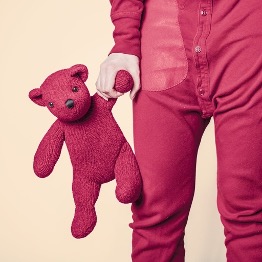} &
        \includegraphics[width=0.17\linewidth]{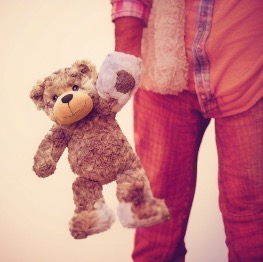} &
        \includegraphics[width=0.17\linewidth]{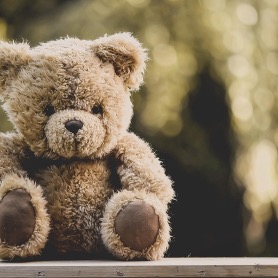} &
        \includegraphics[width=0.17\linewidth]{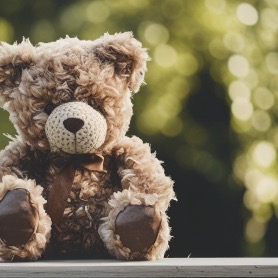} \\[-4.8ex]

        \includegraphics[width=0.17\linewidth]{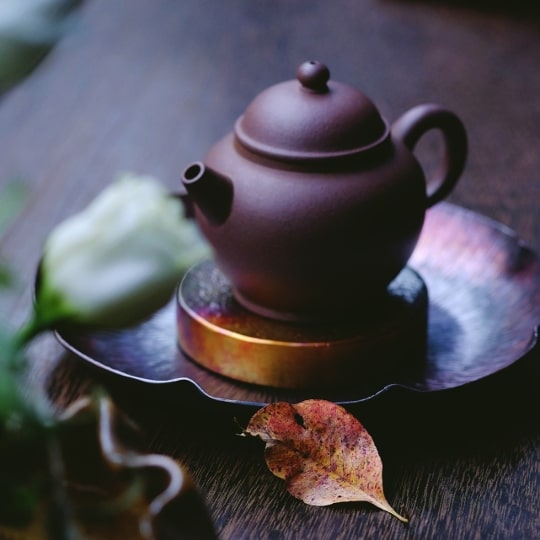} &
        \raisebox{36pt}{$\rightarrow$} &
        \includegraphics[width=0.17\linewidth]{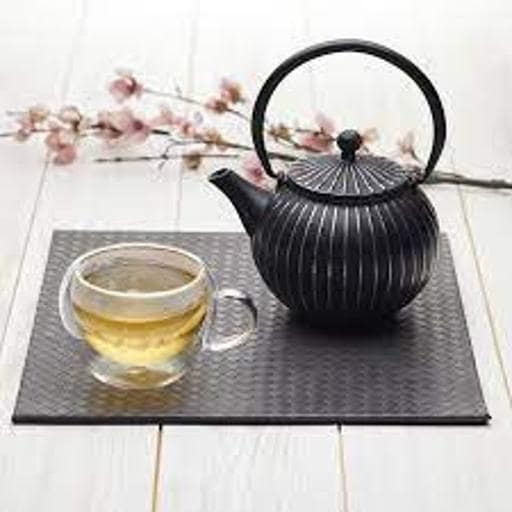} &
        \includegraphics[width=0.17\linewidth]{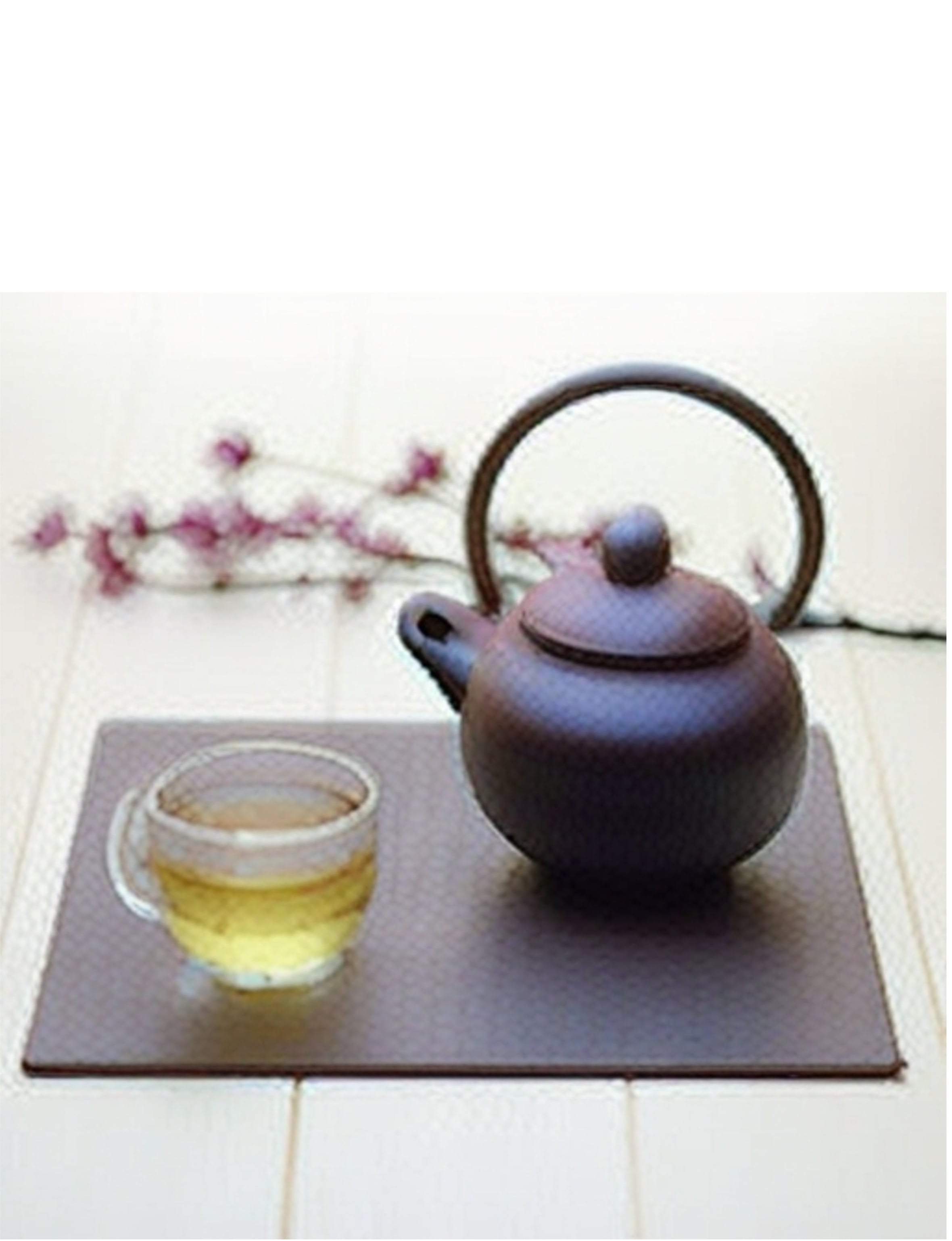} &
        \includegraphics[width=0.17\linewidth]{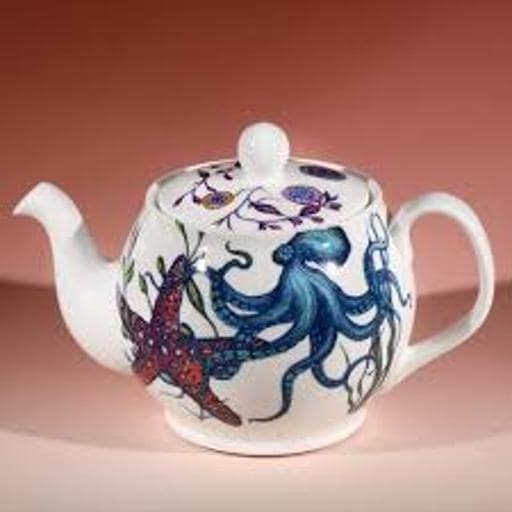} &
        \includegraphics[width=0.17\linewidth]{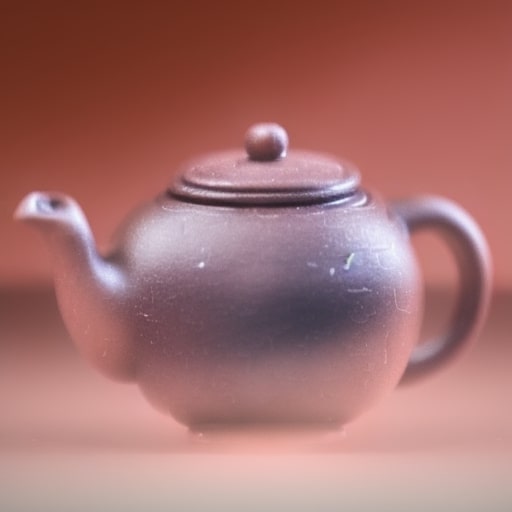} \\

        \includegraphics[width=0.17\linewidth]{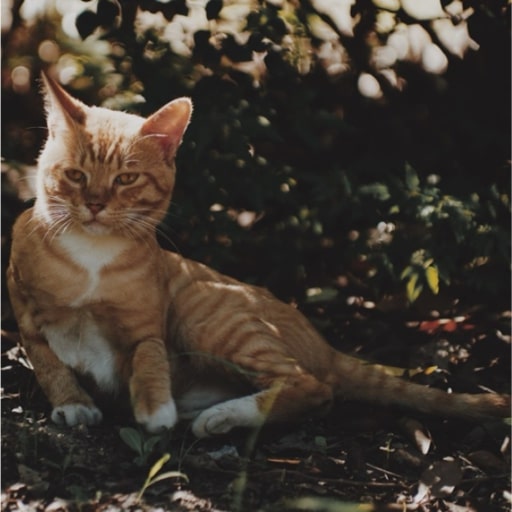} &
        \raisebox{36pt}{$\rightarrow$} &
        \includegraphics[width=0.17\linewidth]{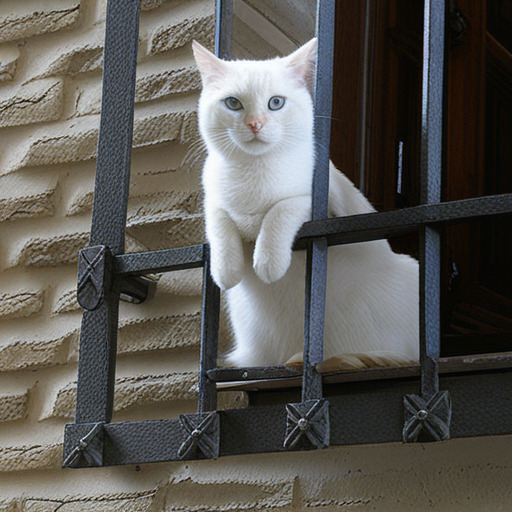} &
        \includegraphics[width=0.17\linewidth]{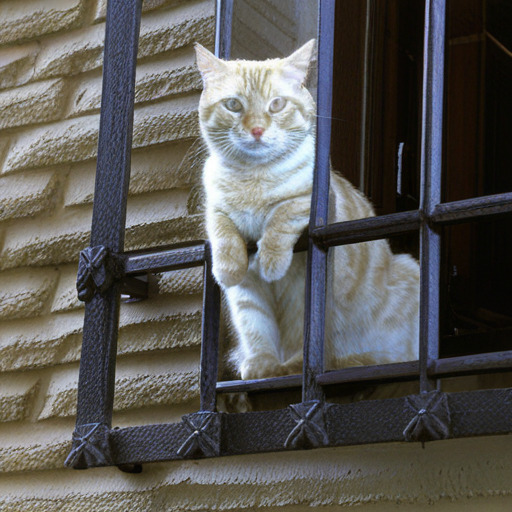} &
        \includegraphics[width=0.17\linewidth]{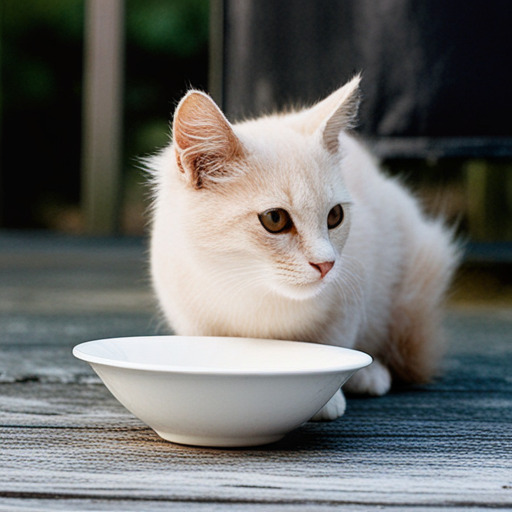} &
        \includegraphics[width=0.17\linewidth]{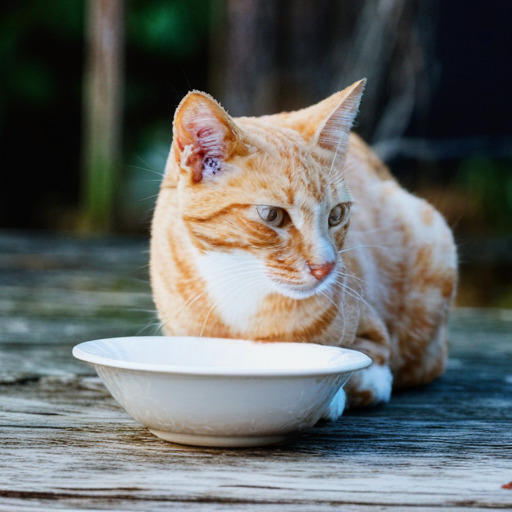} \\

        \includegraphics[width=0.17\linewidth]{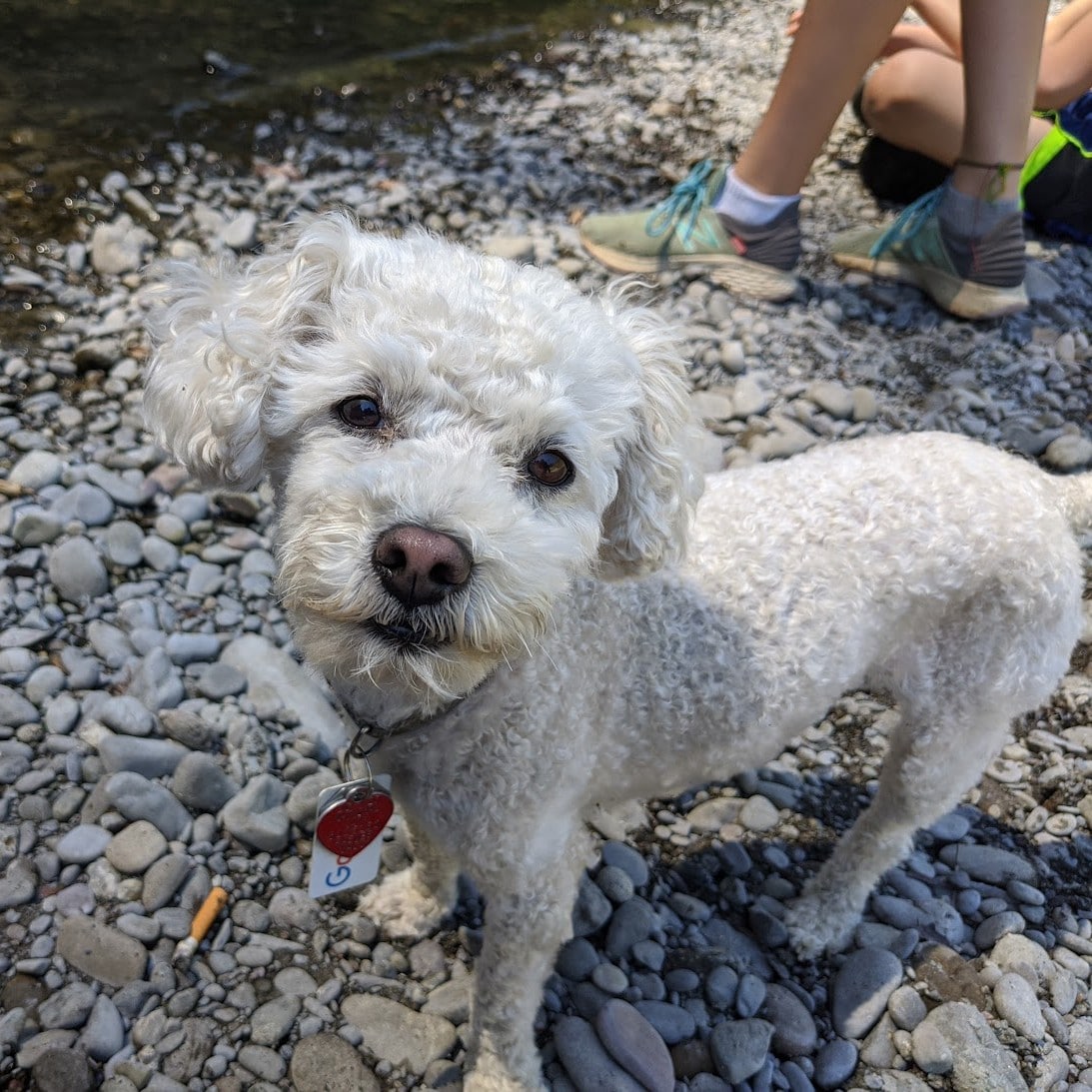} &
        \raisebox{36pt}{$\rightarrow$} &
        \includegraphics[width=0.17\linewidth]{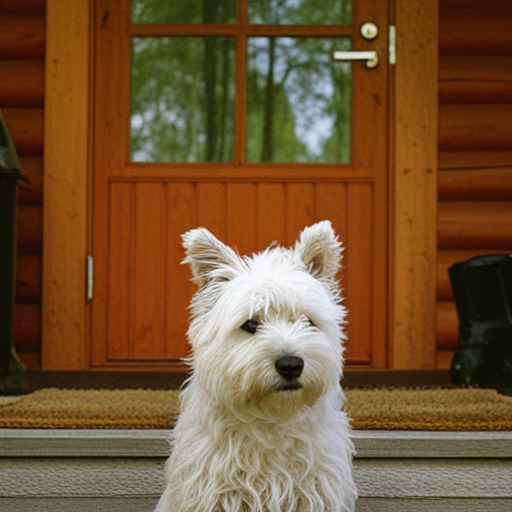} &
        \includegraphics[width=0.17\linewidth]{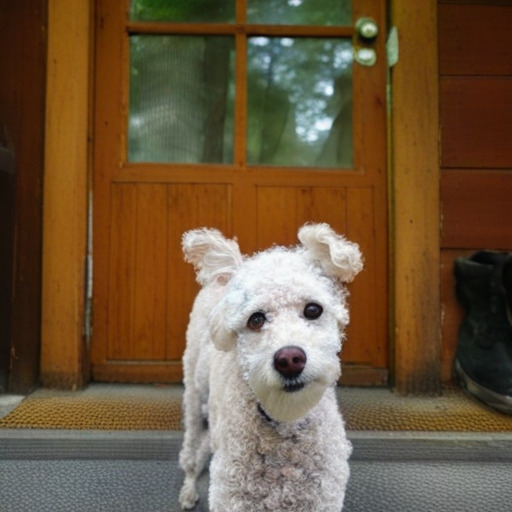} &
        \includegraphics[width=0.17\linewidth]{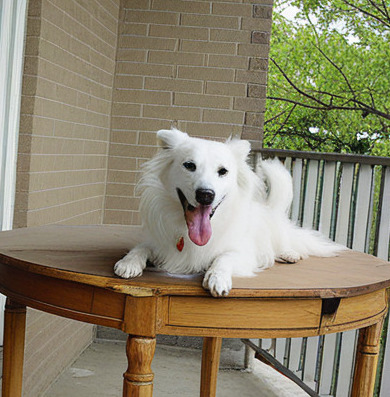} &
        \includegraphics[width=0.17\linewidth]{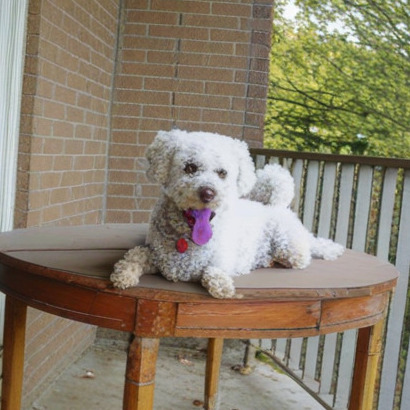} \\

        \includegraphics[width=0.17\linewidth]{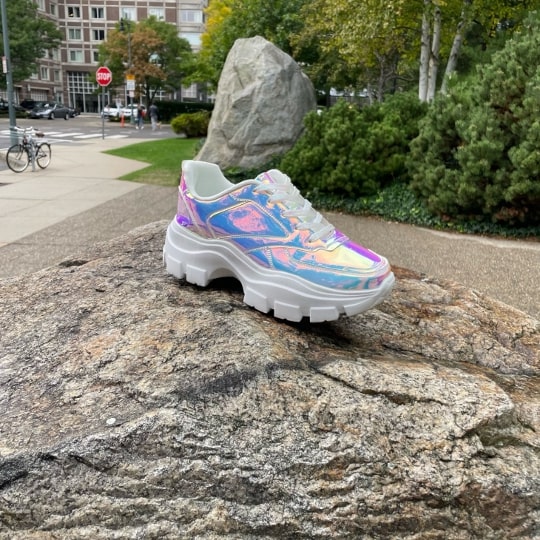} &
        \raisebox{36pt}{$\rightarrow$} &
        \includegraphics[width=0.17\linewidth]{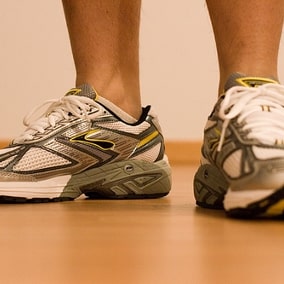} &
        \includegraphics[width=0.17\linewidth]{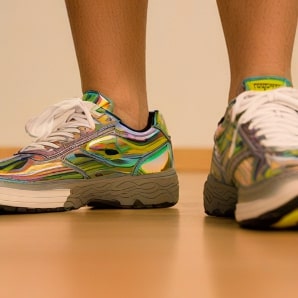} &
        \includegraphics[width=0.17\linewidth]{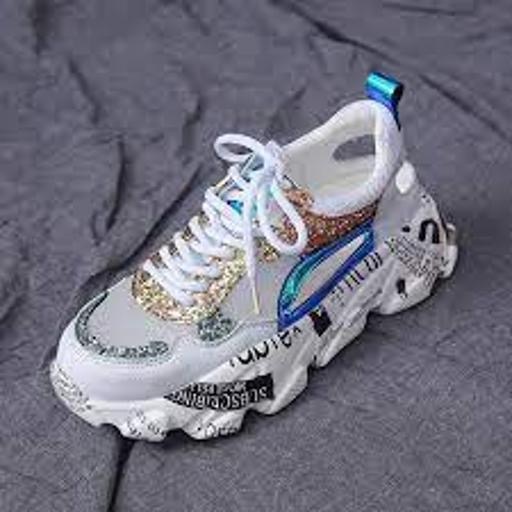} &
        \includegraphics[width=0.17\linewidth]{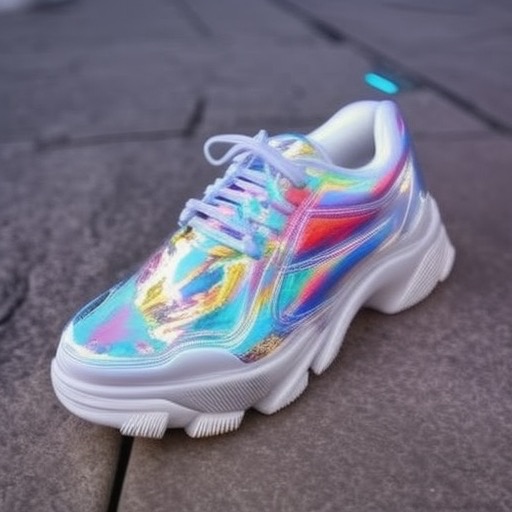} \\

        \includegraphics[width=0.17\linewidth]{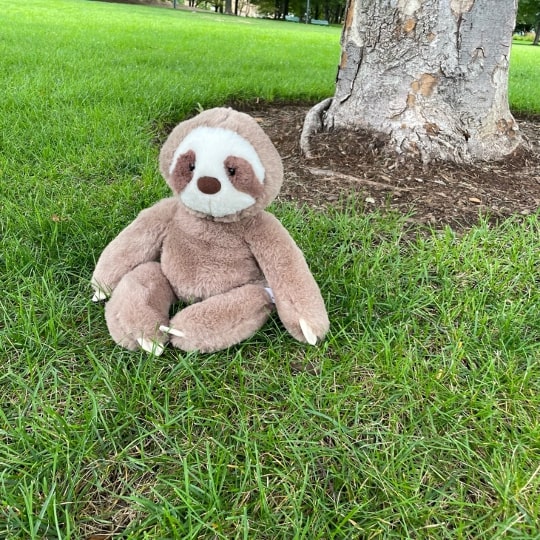} &
        \raisebox{36pt}{$\rightarrow$} &
        \includegraphics[width=0.17\linewidth]{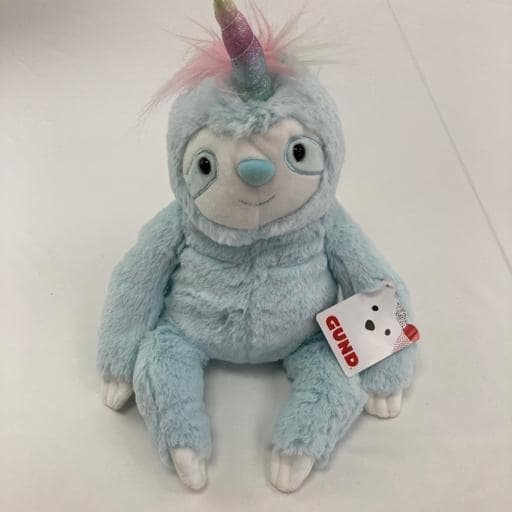} &
        \includegraphics[width=0.17\linewidth]{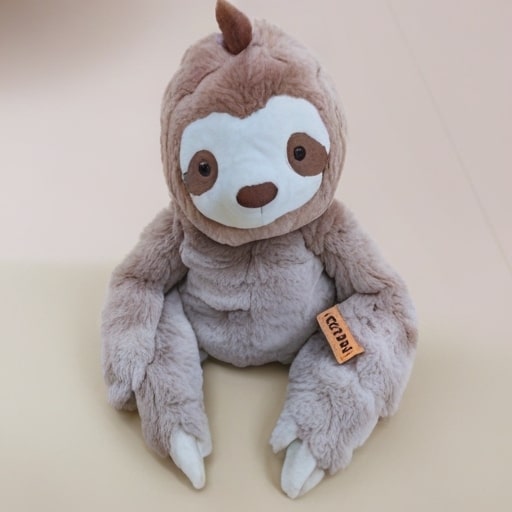} &
        \includegraphics[width=0.17\linewidth]{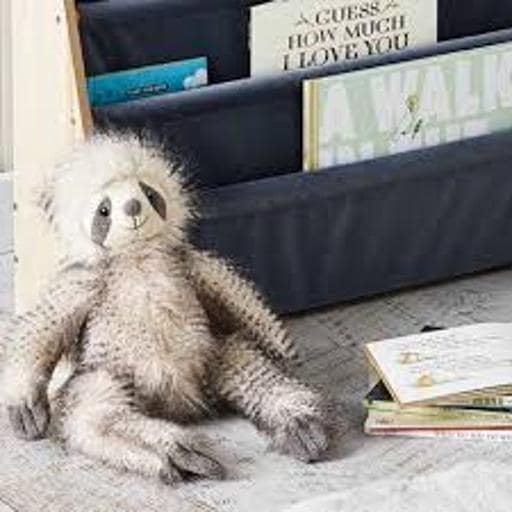} &
        \includegraphics[width=0.17\linewidth]{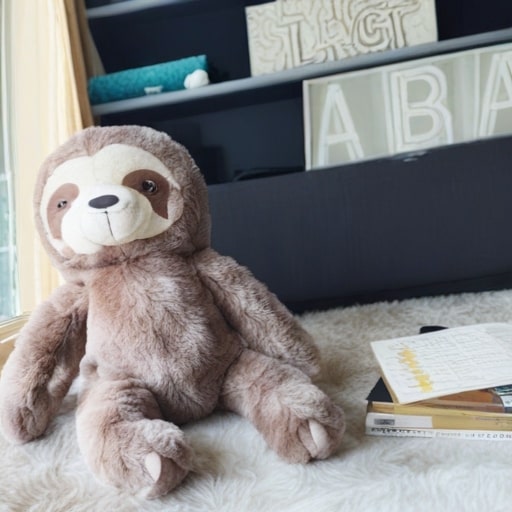} \\
        
    \end{tabular}
    }
    \caption{
        More qualitative results on Subject Driven Image Editing.
    }
    \label{fig:more_editing}
\end{figure*}

\begin{figure*}[t!]
    \centering
    \setlength{\tabcolsep}{1pt}
    \scriptsize{
    \large
    \begin{tabular}{cccc cccccccc}
        & & \text{\small Subject Image} & & & & & & & \\

        \raisebox{18pt}{\rotatebox[origin=t]{90}{\small }} &
        \raisebox{18pt}{\rotatebox[origin=t]{90}{\small ... in the beach}} &
        \includegraphics[width=0.10\linewidth]{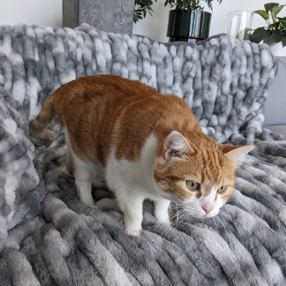} &
        \raisebox{23pt}{$\rightarrow$} &
        \includegraphics[width=0.10\linewidth]{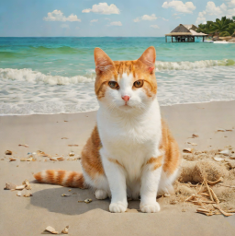} &
        \includegraphics[width=0.10\linewidth]{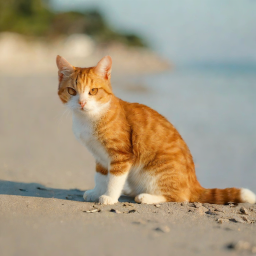} &
        \includegraphics[width=0.10\linewidth]{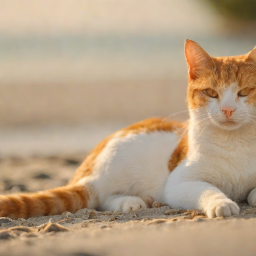} &
        \includegraphics[width=0.10\linewidth]{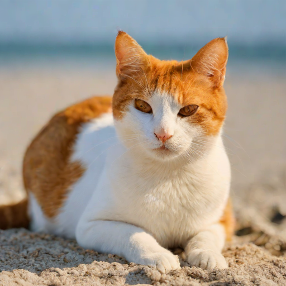} &
        \includegraphics[width=0.10\linewidth]{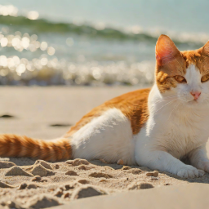} &
        \includegraphics[width=0.10\linewidth]{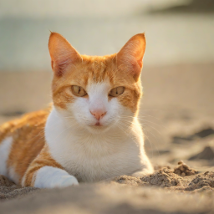} &
        \includegraphics[width=0.10\linewidth]{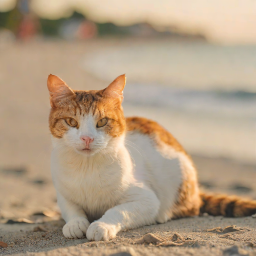} &
        \includegraphics[width=0.10\linewidth]{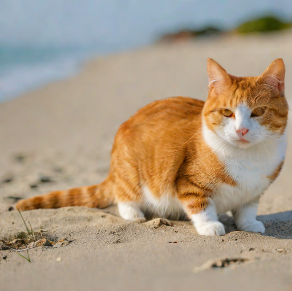} \\

        \raisebox{18pt}{\rotatebox[origin=t]{90}{\small }} &
        \raisebox{18pt}{\rotatebox[origin=t]{90}{\small ... sleeping}} &
        \includegraphics[width=0.10\linewidth]{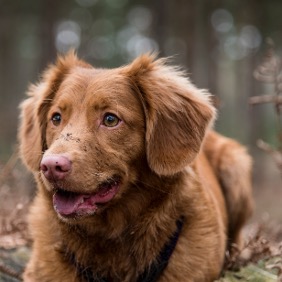} &
        \raisebox{23pt}{$\rightarrow$} &
        \includegraphics[width=0.10\linewidth]{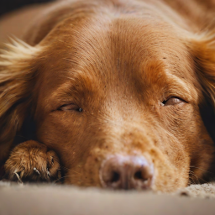} &
        \includegraphics[width=0.10\linewidth]{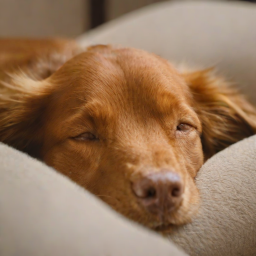} &
        \includegraphics[width=0.10\linewidth]{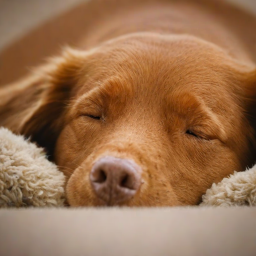} &
        \includegraphics[width=0.10\linewidth]{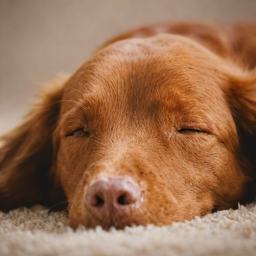} &
        \includegraphics[width=0.10\linewidth]{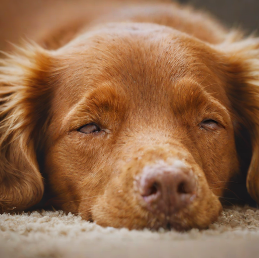} &
        \includegraphics[width=0.10\linewidth]{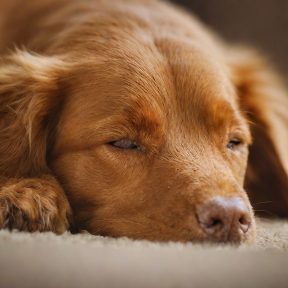} &
        \includegraphics[width=0.10\linewidth]{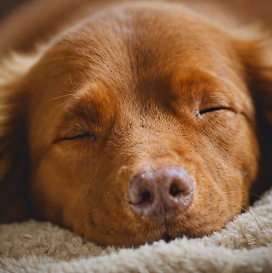} &
        \includegraphics[width=0.10\linewidth]{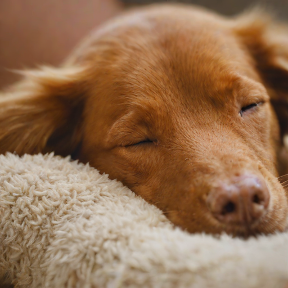} \\

        \raisebox{18pt}{\rotatebox[origin=t]{90}{\small ... on a}} &
        \raisebox{18pt}{\rotatebox[origin=t]{90}{\small glass table}} &
        \includegraphics[width=0.10\linewidth]{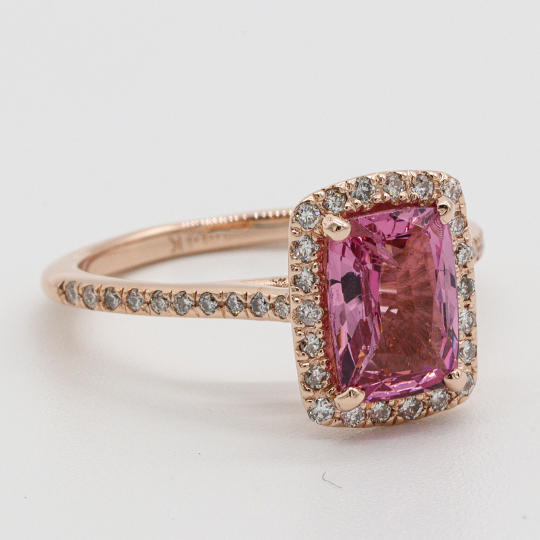} &
        \raisebox{23pt}{$\rightarrow$} &
        \includegraphics[width=0.10\linewidth]{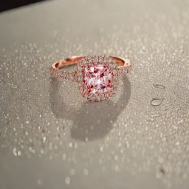} &
        \includegraphics[width=0.10\linewidth]{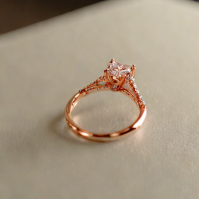} &
        \includegraphics[width=0.10\linewidth]{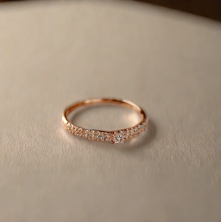} &
        \includegraphics[width=0.10\linewidth]{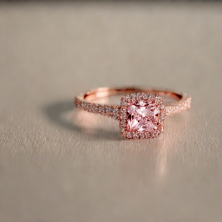} &
        \includegraphics[width=0.10\linewidth]{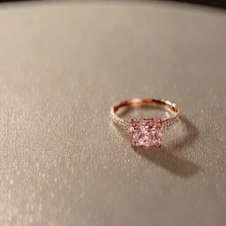} &
        \includegraphics[width=0.10\linewidth]{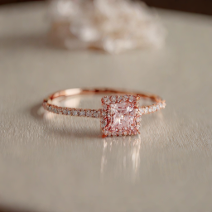} &
        \includegraphics[width=0.10\linewidth]{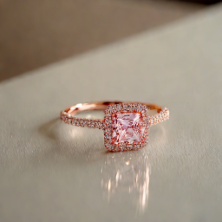} &
        \includegraphics[width=0.10\linewidth]{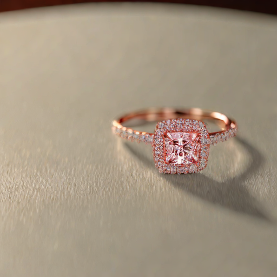} \\

        \raisebox{18pt}{\rotatebox[origin=t]{90}{\small }} &
        \raisebox{18pt}{\rotatebox[origin=t]{90}{\small ... on a bed}} &
        \includegraphics[width=0.10\linewidth]{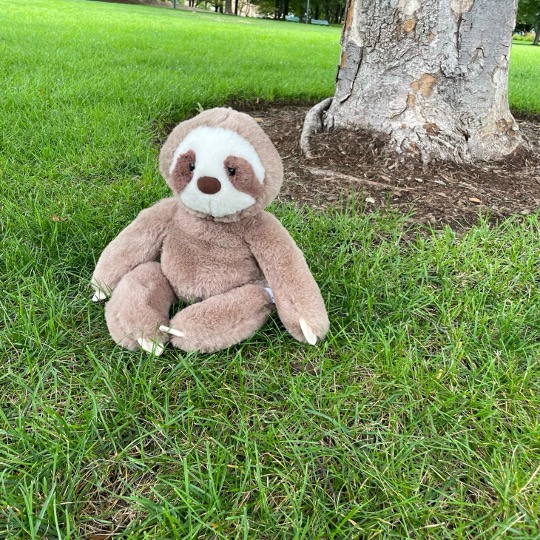} &
        \raisebox{23pt}{$\rightarrow$} &
        \includegraphics[width=0.10\linewidth]{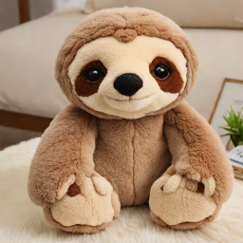} &
        \includegraphics[width=0.10\linewidth]{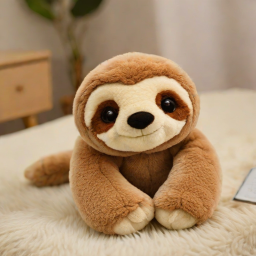} &
        \includegraphics[width=0.10\linewidth]{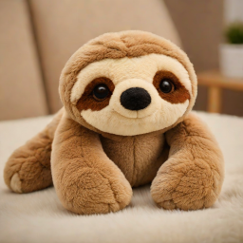} &
        \includegraphics[width=0.10\linewidth]{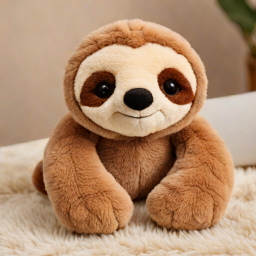} &
        \includegraphics[width=0.10\linewidth]{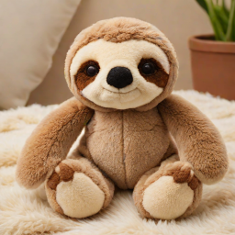} &
        \includegraphics[width=0.10\linewidth]{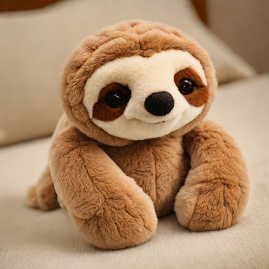} &
        \includegraphics[width=0.10\linewidth]{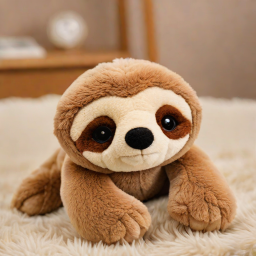} &
        \includegraphics[width=0.10\linewidth]{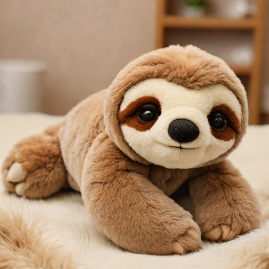} \\

        & & & & \text{\small 10} & \text{\small 20} & \text{\small 30} & \text{\small 35} & \text{\small 42} & \text{\small 50} & \text{\small 100} & \text{\small 120} \\

        & & & & \multicolumn{8}{c}{\small seed value}
        
        \end{tabular}
    }
    \caption{
        We show the stability of our method across eight seeds for Subject Driven Image Generation.
    }
    \label{fig:seed_stability_gen}
\end{figure*}

\begin{figure*}[t!]
    \centering
    \setlength{\tabcolsep}{1pt}
    \scriptsize{
    \large
    \begin{tabular}{ccc cccccccc}
        \text{\small Subject Image} & \text{\small Input Image} & & & & & & & \\

        \includegraphics[width=0.09\linewidth]{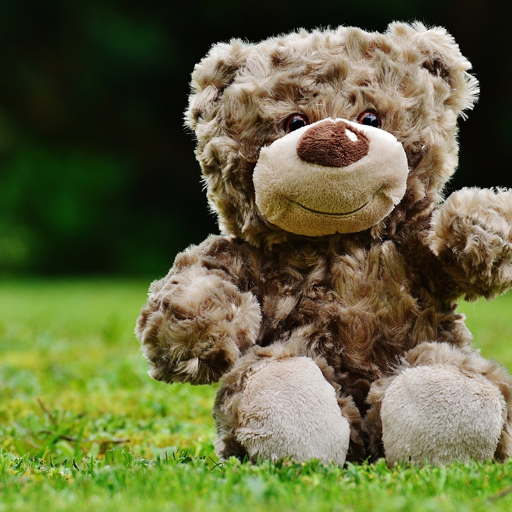} &
        \includegraphics[width=0.09\linewidth]{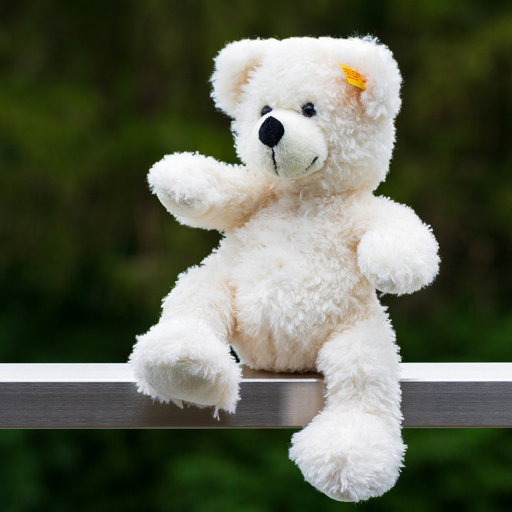} &
        \raisebox{23pt}{$\rightarrow$} &
        \includegraphics[width=0.09\linewidth]{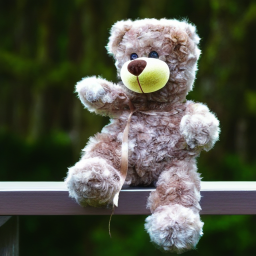} &
        \includegraphics[width=0.09\linewidth]{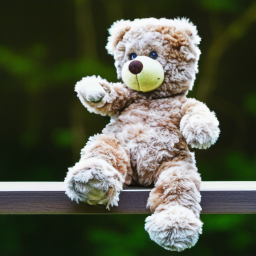} &
        \includegraphics[width=0.09\linewidth]{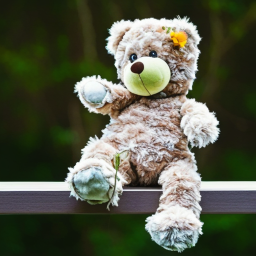} &
        \includegraphics[width=0.09\linewidth]{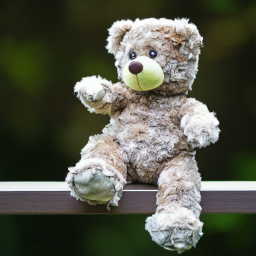} &
        \includegraphics[width=0.09\linewidth]{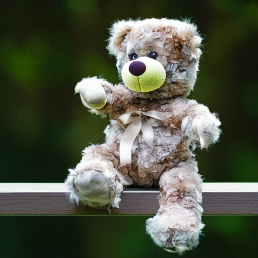} &
        \includegraphics[width=0.09\linewidth]{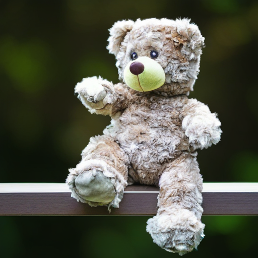} &
        \includegraphics[width=0.09\linewidth]{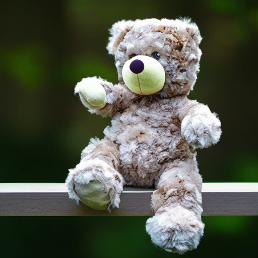} &
        \includegraphics[width=0.09\linewidth]{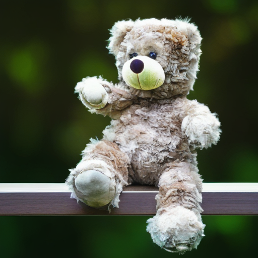} \\

        \includegraphics[width=0.09\linewidth]{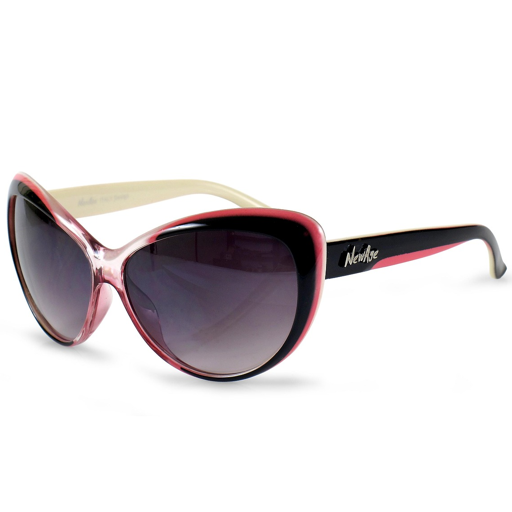} &
        \includegraphics[width=0.09\linewidth]{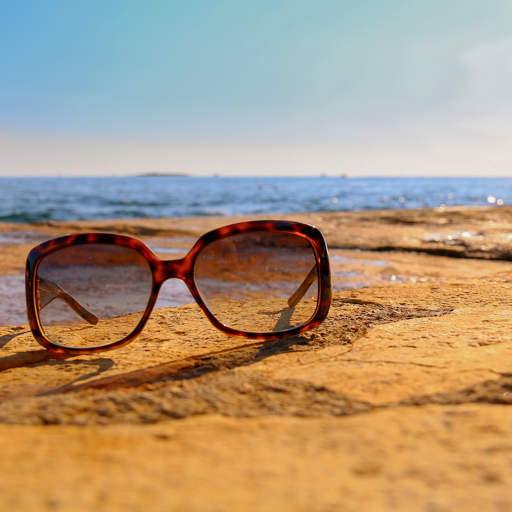} &
        \raisebox{23pt}{$\rightarrow$} &
        \includegraphics[width=0.09\linewidth]{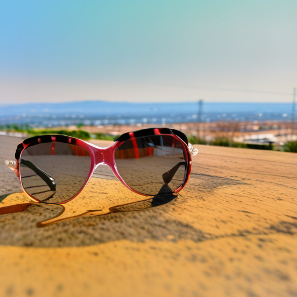} &
        \includegraphics[width=0.09\linewidth]{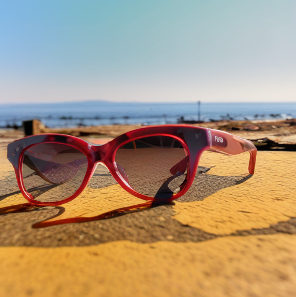} &
        \includegraphics[width=0.09\linewidth]{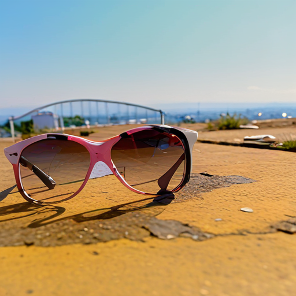} &
        \includegraphics[width=0.09\linewidth]{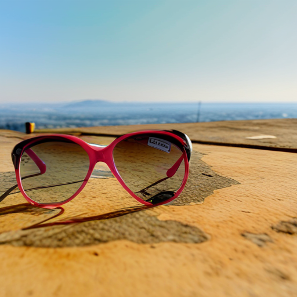} &
        \includegraphics[width=0.09\linewidth]{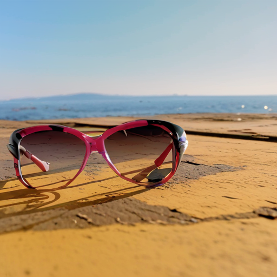} &
        \includegraphics[width=0.09\linewidth]{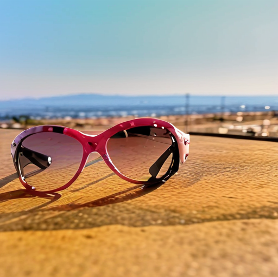} &
        \includegraphics[width=0.09\linewidth]{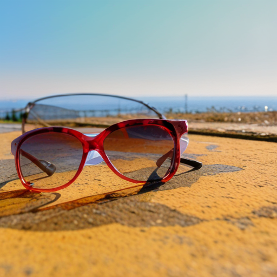} &
        \includegraphics[width=0.09\linewidth]{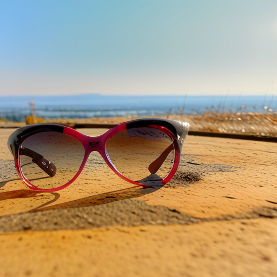} \\

        \includegraphics[width=0.09\linewidth]{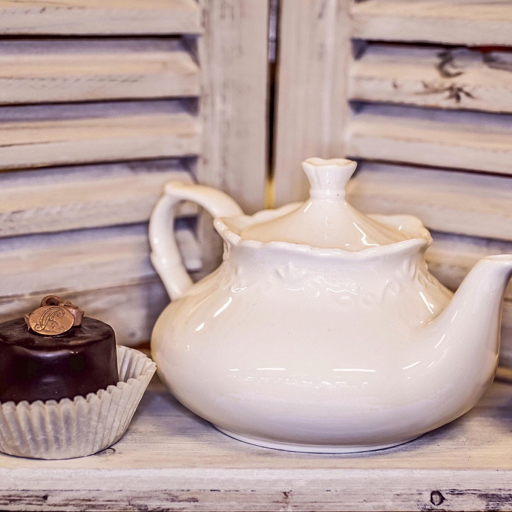} &
        \includegraphics[width=0.09\linewidth]{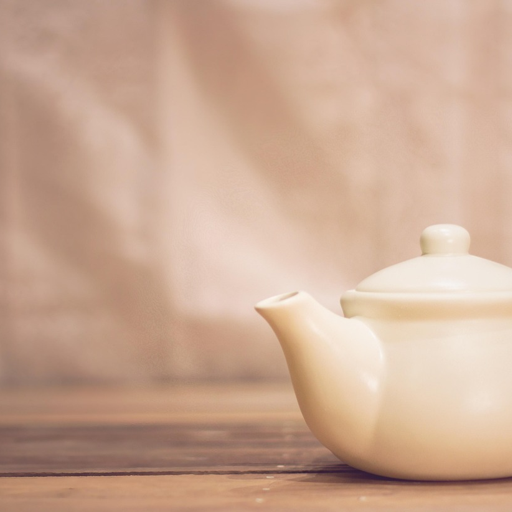} &
        \raisebox{23pt}{$\rightarrow$} &
        \includegraphics[width=0.09\linewidth]{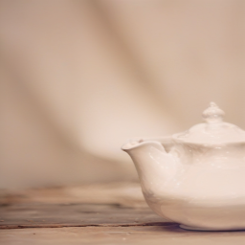} &
        \includegraphics[width=0.09\linewidth]{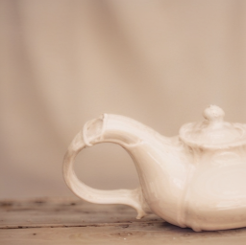} &
        \includegraphics[width=0.09\linewidth]{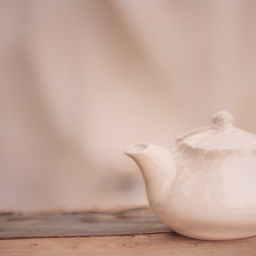} &
        \includegraphics[width=0.09\linewidth]{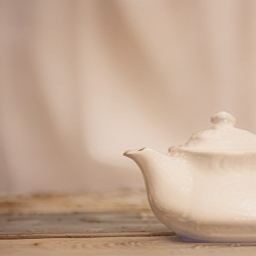} &
        \includegraphics[width=0.09\linewidth]{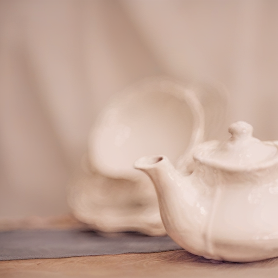} &
        \includegraphics[width=0.09\linewidth]{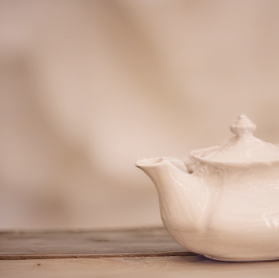} &
        \includegraphics[width=0.09\linewidth]{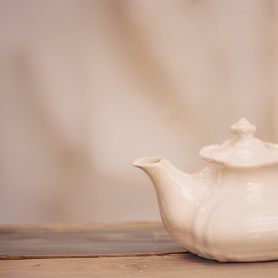} &
        \includegraphics[width=0.09\linewidth]{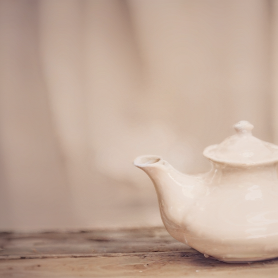} \\

        \includegraphics[width=0.09\linewidth]{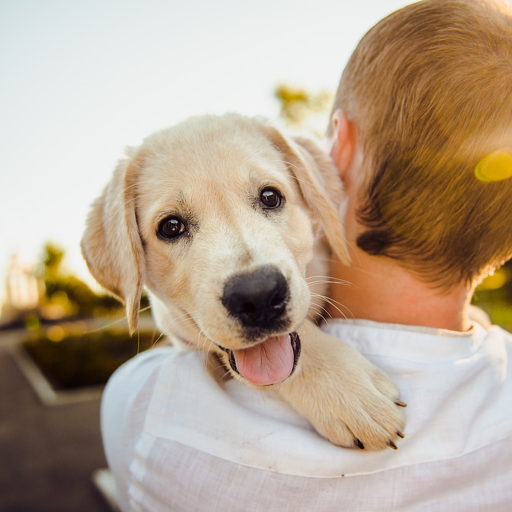} &
        \includegraphics[width=0.09\linewidth]{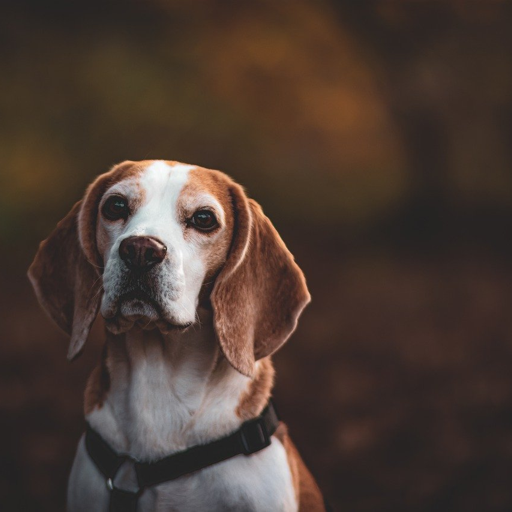} &
        \raisebox{23pt}{$\rightarrow$} &
        \includegraphics[width=0.09\linewidth]{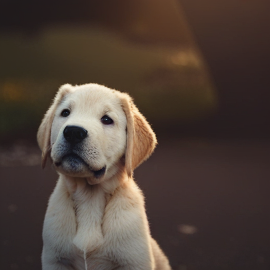} &
        \includegraphics[width=0.09\linewidth]{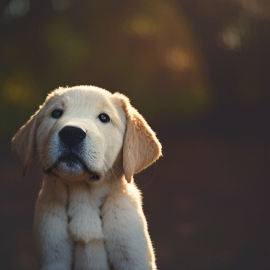} &
        \includegraphics[width=0.09\linewidth]{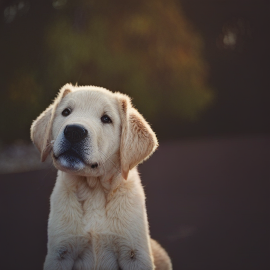} &
        \includegraphics[width=0.09\linewidth]{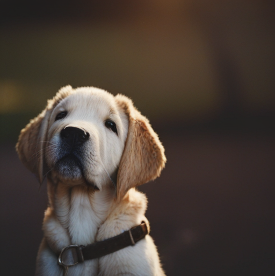} &
        \includegraphics[width=0.09\linewidth]{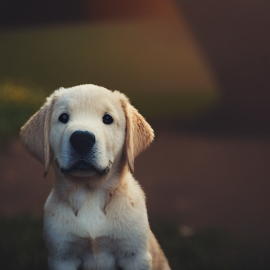} &
        \includegraphics[width=0.09\linewidth]{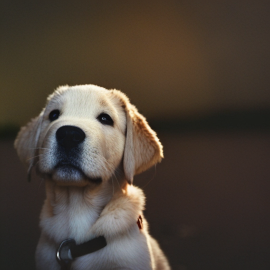} &
        \includegraphics[width=0.09\linewidth]{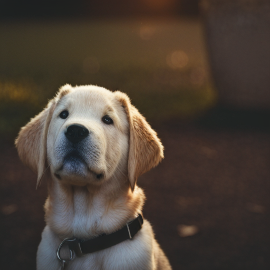} &
        \includegraphics[width=0.09\linewidth]{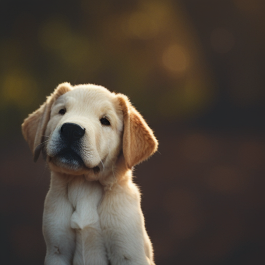} \\
        
        & & & \text{\small 10} & \text{\small 20} & \text{\small 30} & \text{\small 35} & \text{\small 42} & \text{\small 50} & \text{\small 100} & \text{\small 120} \\

        & & & \multicolumn{8}{c}{\small seed value}
        
        \end{tabular}
    }
    \caption{
        We show the stability of our method across eight seeds for Subject Driven Image Editing.
    }
    \label{fig:seed_stability_edit}
\end{figure*}

\end{document}